\documentclass[conference,compsoc]{IEEEtran}

\usepackage{amssymb,epsfig}
\usepackage{graphicx}
\usepackage{subcaption}
\usepackage{amsmath}
\usepackage{amsfonts}
\usepackage{caption}
\usepackage{tabularx}
\usepackage{multirow}
\usepackage{gensymb}
\usepackage{bm}
\usepackage{color}

\DeclareMathOperator*{\argmin}{arg\,min}

\ifCLASSOPTIONcompsoc
  \usepackage[nocompress]{cite}
\else
  \usepackage{cite}
\fi

\ifCLASSINFOpdf
\else
\fi

\begin{document}


\title{A Survey on Deep Learning for Localization and Mapping: \\ Towards the Age of Spatial Machine Intelligence}


\author{Changhao Chen, Bing Wang, Chris Xiaoxuan Lu, Niki Trigoni and Andrew Markham\\
Department of Computer Science, University of Oxford\\
\IEEEcompsocitemizethanks{\IEEEcompsocthanksitem Corresponding author: Changhao Chen (changhao.chen@cs.ox.ac.uk)
}
\IEEEcompsocitemizethanks{\IEEEcompsocthanksitem A project website that updates additional material and extended lists of references, can be found at https://github.com/changhao-chen/deep-learning-localization-mapping.
}
}

\maketitle

\begin{abstract}

Deep learning based localization and mapping has recently attracted significant attention. Instead of creating hand-designed algorithms through exploitation of physical models or geometric theories, deep learning based solutions provide an alternative to solve the problem in a data-driven way.
Benefiting from ever-increasing volumes of data and computational power, these methods are fast evolving into a new area that offers accurate and robust systems to track motion and estimate scenes and their structure for real-world applications.
In this work, we provide a comprehensive survey, and propose a new taxonomy for localization and mapping using deep learning. We also discuss the limitations of current models, and indicate possible future directions. 
A wide range of topics are covered, from learning odometry estimation, mapping, to global localization and simultaneous localization and mapping (SLAM). 
We revisit the problem of perceiving self-motion and scene understanding with on-board sensors, and show how to solve it by integrating these modules into a prospective spatial machine intelligence system (SMIS). It is our hope that this work can connect emerging works from robotics, computer vision and machine learning communities, and serve as a guide for future researchers to apply deep learning to tackle localization and mapping problems.

\end{abstract}


\IEEEpeerreviewmaketitle

\section{Introduction}
Localization and mapping is a fundamental need for human and mobile agents.
As a motivating example, humans are able to perceive their self-motion and environment via multimodal sensory perception, and rely on this awareness to locate and navigate themselves in a complex three-dimensional space \cite{Fetsch2009}. This ability is part of human spatial ability.
Furthermore, the ability to perceive self-motion and their surroundings plays a vital role in developing cognition, and motor control \cite{Cullen2012}. 
In a similar vein, artificial agents or robots should also be able to perceive the environment and estimate their system states using on-board sensors. 
These agents could be any form of robot, e.g. self-driving vehicles, delivery drones or home service robots, sensing their surroundings and autonomously making decisions \cite{Sunderhauf2018}. Equivalently, as emerging Augmented Reality (AR) and Virtual Reality (VR) technologies interweave cyber space and physical environments, the ability of machines to be perceptually aware underpins seamless human-machine interaction. Further applications also include mobile and wearable devices, such as smartphones, wristbands or Internet-of-Things (IoT) devices, providing users with a wide range of location-based-services, ranging from pedestrian navigation \cite{Harle2013}, to sports/activity monitoring \cite{Gowda2017}, to animal tracking \cite{wijers2019caracal}, or emergency response \cite{Dhekne2019} for first-responders. 

   \begin{figure}
   	\centering
        \includegraphics[width=0.48\textwidth]{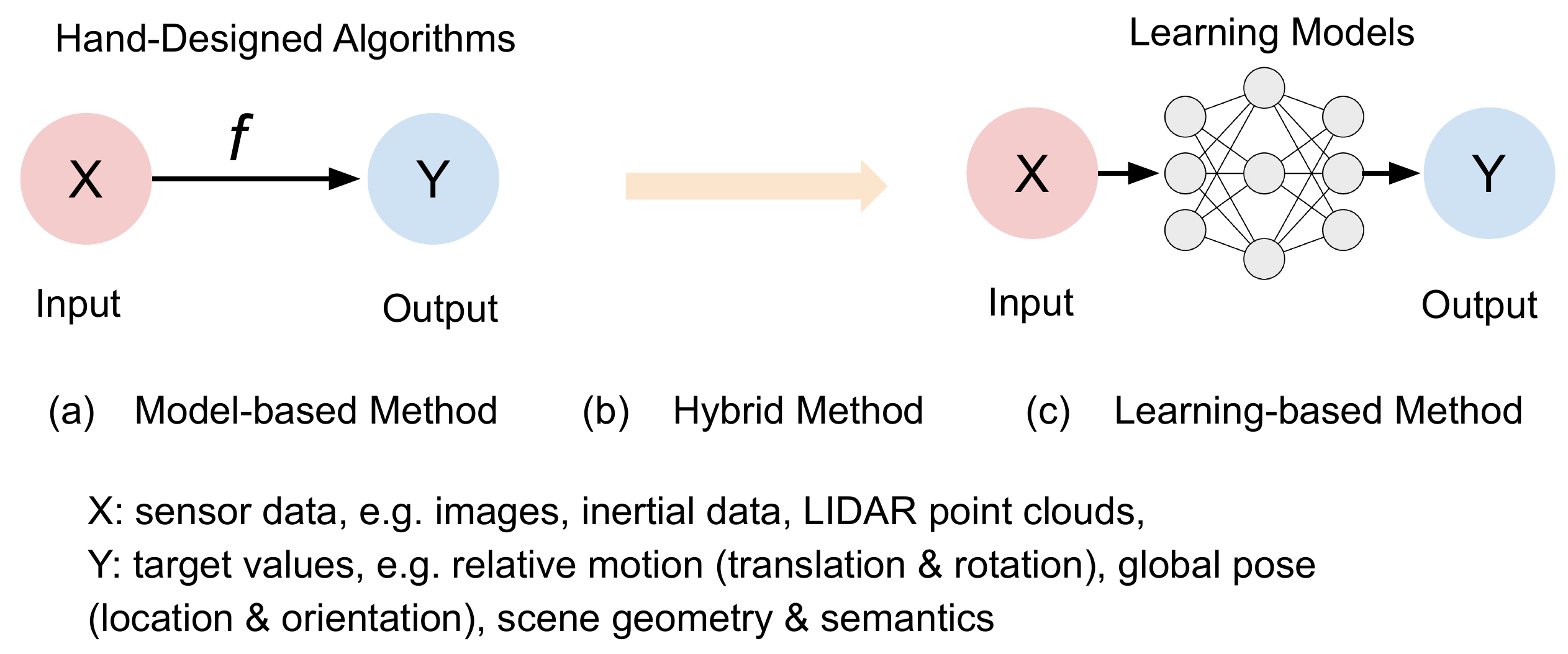}
        \caption{
        A spatial machine intelligence system exploits on-board sensors to perceive self-motion, global pose, scene geometry and semantics.
        (a) Conventional model based solutions build hand-designed algorithms to convert input sensor data to target values. (c) Data-driven solutions exploit learning models to construct this mapping function. (b) Hybrid approaches combine both hand-crafted algorithms and learning models. This survey discusses (b) and (c). }
        \label{fig: overview}
    \end{figure} 
    
    \begin{figure*}
        \centering
        \includegraphics[width=0.95\textwidth]{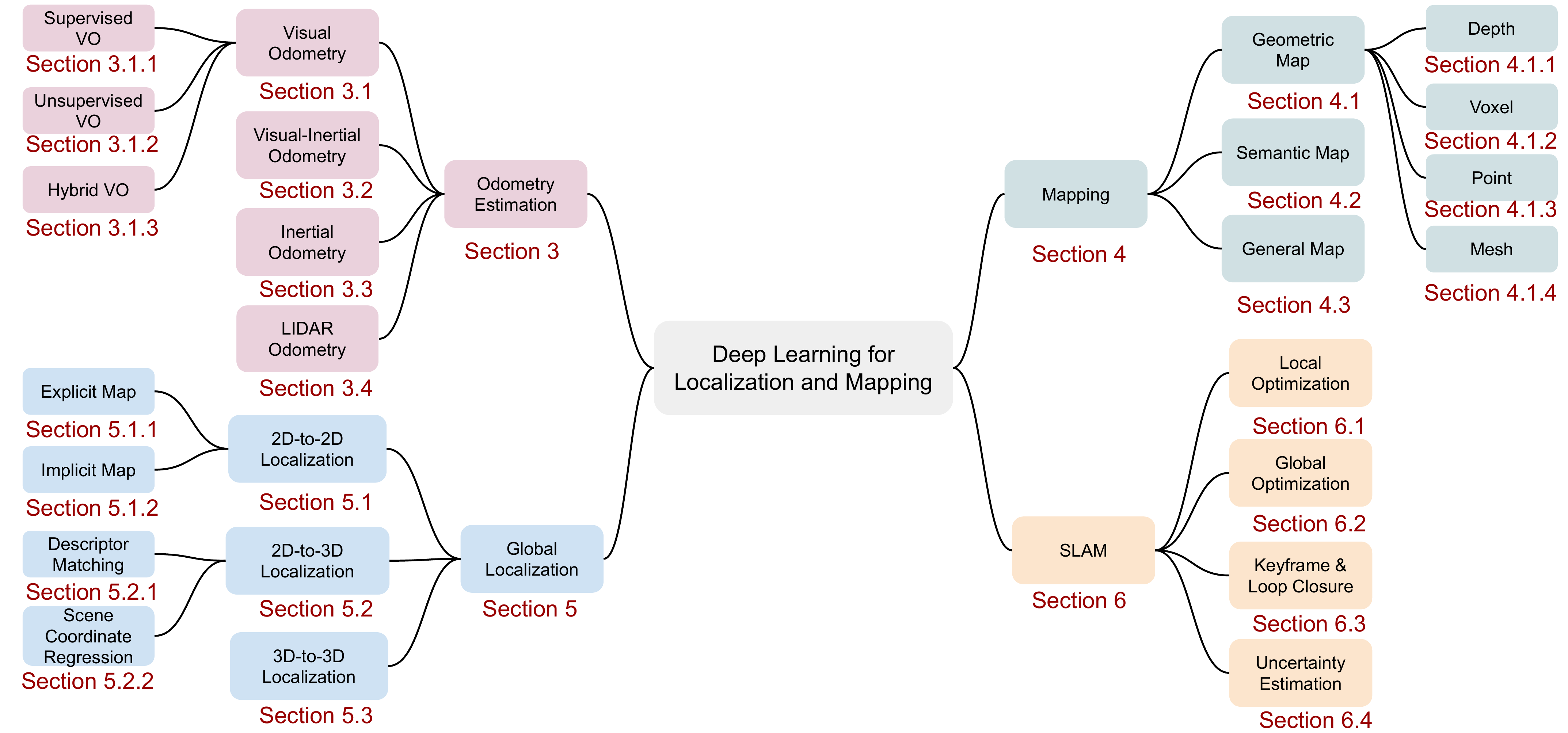}
        \caption{A taxonomy of existing works on deep learning for localization and mapping.}
        \label{fig: taxonomy}
    \end{figure*}

Enabling a high level of autonomy for these and other digital agents requires precise and robust localization, and incrementally building and maintaining a world model, with the capability to continuously process new information and adapt to various scenarios. Such a quest is termed as `\textit{Spatial Machine Intelligence System (SMIS)}' in our work or recently as Spatial AI in \cite{Davison2018}.
In this work, broadly, \textit{localization} refers to the ability to obtain the internal system states of robot motion, including locations, orientations and velocities, whilst \textit{mapping} indicates the capacity to perceive external environmental states and capture the surroundings, including the geometry, appearance and semantics of a 2D or 3D scene. These components can act individually to sense the internal or external states respectively, or jointly as in simultaneous localization and mapping (SLAM) to track pose and build a consistent environmental model in a global frame.

\subsection{Why to Study Deep Learning for Localization and Mapping}
The problems of localization and mapping have been studied for decades, with a variety of intricate hand-designed models and algorithms being developed, for example, odometry estimation (including visual odometry \cite{Nister2004,Engel2013,Forster2014}, visual-inertial odometry \cite{Li2013b,Leutenegger2015,Forster2017,Qin2018} and LIDAR odometry \cite{Zhang2010}), image-based localization\cite{zhang2006image,sattler2011fast}, place recognition\cite{lowry2015visual}, SLAM\cite{Davison2007,Engel2013,Montiel2015}, and structure from motion (SfM)\cite{longuet1981computer,wu2013towards}.
Under ideal conditions, these sensors and models are capable of accurately estimating system states without time bound and across different environments.
However, in reality, imperfect sensor measurements, inaccurate system modelling, complex environmental dynamics and unrealistic constraints impact both the accuracy and reliability of hand-crated systems. 

The limitations of model based solutions, together with recent advances in machine learning, especially deep learning, have motivated researchers to consider data-driven (learning) methods as an alternative to solve problem.
Figure \ref{fig: overview} summarizes the relation between input sensor data (e.g. visual, inertial, LIDAR data or other sensors) and output target values (e.g. location, orientation, scene geometry or semantics) as a mapping function.
Conventional model-based solutions are achieved by hand-designing algorithms and calibrating to a particular application domain, while the learning based approaches construct this mapping function by learned knowledge.
The advantages of learning based methods are three-fold:

First of all, learning methods can leverage highly expressive deep neural network as an universal approximator, and automatically discover features relevant to task. This property enables learned models to be resilience to circumstances, such as featureless areas, dynamic lightning conditions, motion blur, accurate camera calibration, which are challenging to model by hand \cite{Sunderhauf2018}. As a representative example, visual odometry has achieved notable improvements in terms of robustness by incorporating data-driven methods in its design \cite{Wang2017,yang2020d3vo}, outperforming the state-of-the-art conventional algorithms. Moreover, learning approaches are able to connect abstract elements with human understandable terms\cite{mccormac2017semanticfusion,ma2017multi}, such as semantics labelling in SLAM, which is hard to describe in a formal mathematical way.

\begin{figure*}
    \centering
    \includegraphics[width=0.95\textwidth]{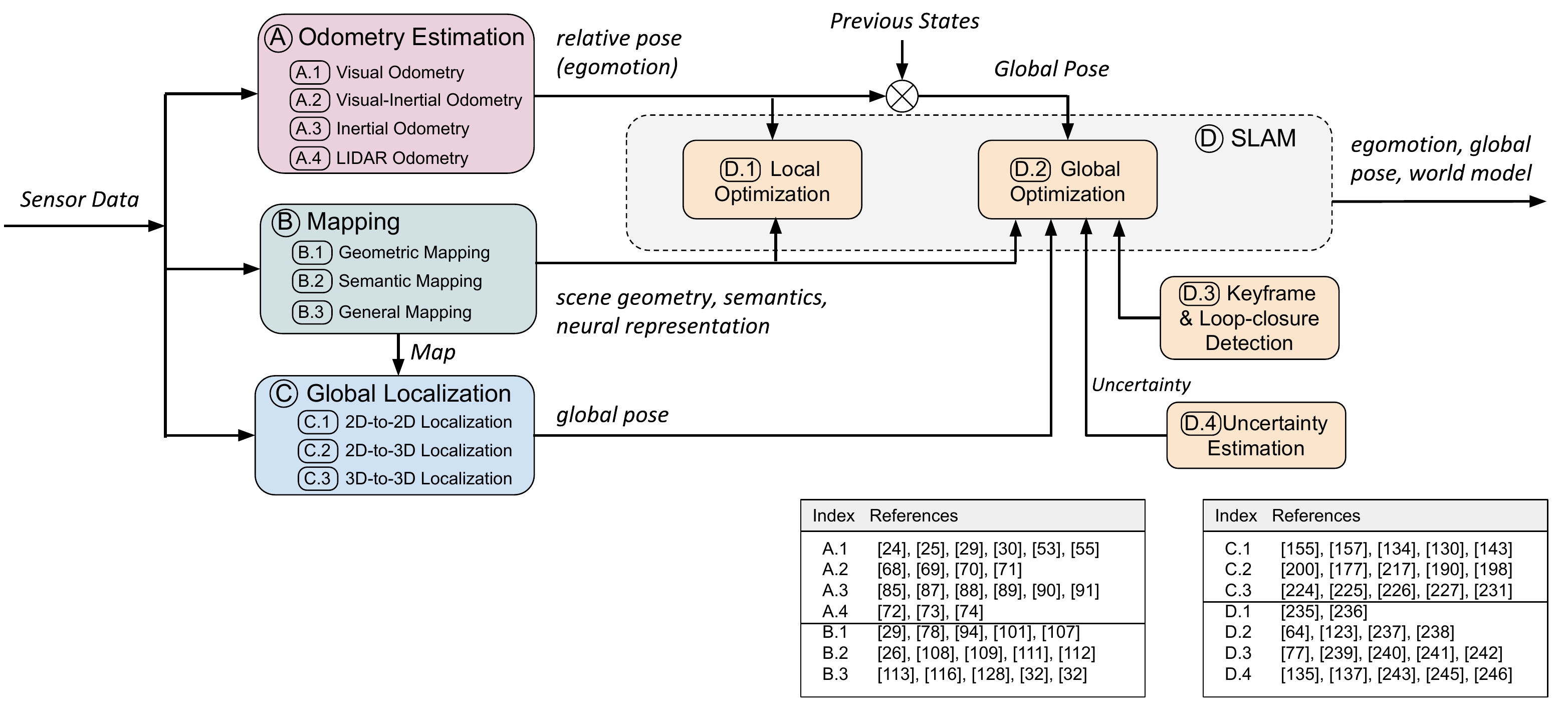}
    \caption{High-level conceptual illustration of a spatial machine intelligence system (i.e. deep learning based localization and mapping). Rounded rectangles represent a function module, while arrow lines connect these modules for data input and output. It is not necessary to include all modules to perform this system.}
    \label{fig: concept figure}
\end{figure*}

Secondly, 
learning methods allow spatial machine intelligence systems to learn from past experience, and actively exploit new information. By building a generic data-driven model, it avoids human effort on specifying the full knowledge about mathematical and physical rules\cite{goodfellow2016deep}, to solve domain specific problem, before being deployed. This ability potentially enables learning machines to automatically discover new computational solutions, further develop themselves and improve their models, within new scenarios or confronting new circumstances. A good example is that by using novel view synthesis as a self-supervision signal, self-motion and depth can be recovered from unlabelled videos\cite{Zhou2017,bian2019unsupervised}. 
In addition, the learned representations can further support high-level tasks, such as path planning\cite{zhu2017target}, and decision making\cite{mirowski2018learning}, by constructing task-driven maps.

The third benefit is its capability of fully exploiting the increasing amount of sensor data and computational power. 
Deep learning or deep neural network has the capacity to scale to large-scale problems. The huge amount of parameters inside a DNN framework are automatically optimized by minimizing a loss function, by training on large datasets through backpropagation and gradient-descent algorithms.  
For example, the recent released GPT-3\cite{brown2020language}, the largest pretrained language model, with incredibly over 175 Billion parameters, 
achieves the state-of-the-art results on a variety of natural language processing (NLP) tasks, even without fine-tuning.
In addition, a variety of large-scale datasets relevant to localization and mapping have been released, for example, in the autonomous vehicles scenarios, \cite{Maddern2016,wang2019apolloscape,sun2019scalability} are with a collection of rich combinations of sensor data, and motion and semantic labels.
This gives us an imagination that it would be possible to exploit the power of data and computation in solving localization and mapping.

However, it must also be pointed out that these learning techniques are reliant on massive datasets to extract statistically meaningful patterns and can struggle to generalize to out-of-set environments. There is lack of model interpretability. Additionally, although highly parallelizable, they are also typically more computationally costly than simpler models. Details of limitations are discussed in Section 7.


\subsection{Comparison with Other Surveys}
There are several survey papers that have extensively discussed model-based localization and mapping approaches. The development of SLAM problem in early decades has been well summarized in \cite{durrant2006simultaneous,bailey2006simultaneous}. The seminal survey \cite{cadena2016past} provides a thorough discussion on existing SLAM work, reviews the history of development and charts several future directions. Although this paper contains a section which briefly discusses deep learning models, it does not overview this field comprehensively, especially due to the explosion of research in this area of the past five years. 
Other SLAM survey papers only focus on individual flavours of SLAM systems, including the probabilistic formulation of SLAM \cite{thrun2005probabilistic}, visual odometry \cite{scaramuzza2011visual}, pose-graph SLAM \cite{grisetti2010tutorial}, and SLAM in dynamic environments \cite{saputra2018visual}. 
We refer readers to these surveys for a better understanding of the conventional model based solutions.
On the other hand, \cite{Sunderhauf2018} has a discussion on the applications of deep learning to robotics research; however, its main focus is not on localization and mapping specifically, but a more general perspective towards the potentials and limits of deep learning in a broad context of robotics, including policy learning, reasoning and planning. 

Notably, although the problem of localization and mapping falls into the key notion of robotics, the incorporation of learning methods progresses in tandem with other research areas such as machine learning, computer vision and even natural language processing. This cross-disciplinary area thus imposes non-trivial difficulty when comprehensively summarizing related works into a survey paper.
To the best of our knowledge, this is the first survey article that thoroughly and extensively covers existing work on deep learning for localization and mapping. 

\subsection{Survey Organization}
The remainder of the paper is organized as follows: Section 2 offers an overview and presents a taxonomy of existing deep learning based localization and mapping; Sections 3, 4, 5, 6 discuss the existing deep learning works on relative motion (odometry) estimation, mapping methods for geometric, semantic and general, global localization, and simultaneous localization and mapping with a focus on SLAM back-ends respectively; Open questions are summarized in Section 7 to discuss the limitations and future prospects of existing work; and finally Section 8 concludes the paper.


\section{Taxonomy of Existing Approaches}

We provide a new taxonomy of existing deep learning approaches, relevant to localization and mapping, to connect the fields of robotics, computer vision and machine learning.
Broadly, they can be categorized into odometry estimation, mapping, global localization and SLAM, as illustrated by the taxonomy shown in Figure \ref{fig: taxonomy}:

1) \textit{Odometry estimation} concerns the calculation of the relative change in pose, in terms of translation and rotation, between two or more frames of sensor data. It continuously tracks self-motion, and is followed by a process to integrate these pose changes with respect to an initial state to derive global pose, in terms of position and orientation. This is widely known as the so-called dead reckoning solution. Odometry estimation can be used in providing pose information and as odometry motion model to assist the feedback loop of robot control.
The key problem is to accurately estimate motion transformations from various sensor measurements.
To this end, deep learning is applied to model the motion dynamics in an end-to-end fashion or extract useful features to support a pre-built system in a hybrid way. 

2) \textit{Mapping} builds and reconstructs a consistent model to describe the surrounding environment.  
Mapping can be used to provide environment information for human operators and high-level robot tasks, constrain the error drifts of odometry estimation, and retrieve the inquiry observation for global localization \cite{cadena2016past}. 
Deep learning is leveraged as an useful tool to discover scene geometry and semantics from high-dimensional raw data for mapping. Deep learning based mapping methods are sub-divided into geometric, semantic, and general mapping, depending on whether the neural network learns the explicit geometry or semantics of a scene, or encodes the scene into an implicit neural representation respectively.

3) \textit{Global localization} retrieves the global pose of mobile agents in a known scene with prior knowledge. This is achieved by matching the inquiry input data with a pre-built 2D or 3D map, other spatial references, or a scene that has been visited before.
It can be leveraged to reduce the pose drift of a dead reckoning system or solve the 'kidnapped robot' problem\cite{thrun2005probabilistic}.  
Deep learning is used to tackle the tricky data association problem that is complicated by the changes in views, illumination, weather and scene dynamics, between the inquiry data and map.

4) \textit{Simultaneous Localisation and Mapping (SLAM)} integrates the aforementioned odometry estimation, global localization and mapping processes as front-ends, and jointly optimizes these modules to boost performance in both localization and mapping. Except these abovementioned modules, several other SLAM modules perform to ensure the consistency of the entire system as follows: \emph{local optimization} ensures the local consistency of camera motion and scene geometry; \emph{global optimization} aims to constrain the drift of global trajectories, and  in a global scale; \emph{keyframe detection} is used in keyframe-based SLAM to enable more efficient inference, while system error drifts can be mitigated by global optimization, once a loop closure is detected by \emph{loop-closure detection}; \emph{uncertainty estimation} provides a metric of belief in the learned poses and mapping, critical to probabilistic sensor fusion and back-end optimization in SLAM systems.

\begin{figure*}
    \centering
    \includegraphics[width=0.95\textwidth]{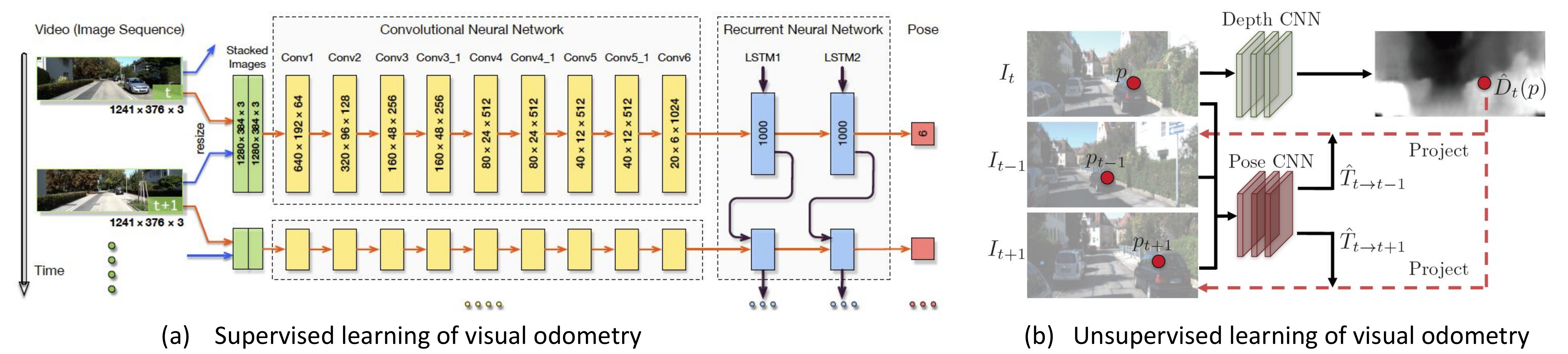}
    \caption{The typical structure of supervised learning of visual odometry, i.e. DeepVO \cite{Wang2017} and unsupervised learning of visual odometry, i.e. SfmLearner \cite{Zhou2017}.}
    \label{fig: vo}
\end{figure*}

Despite the different design goals of individual components, the above components can be integrated into a spatial machine intelligence system (SMIS) to solve real-world challenges, allowing for robust operation, and long-term autonomy in the wild. 
A conceptual figure of such an integrated deep-learning based localization and mapping system is indicated in Figure \ref{fig: concept figure}, showing the relationship of these components. 
In the following sections, we will discuss these components in details.

\section{Odometry Estimation}
We begin with odometry estimation, which continuously tracks camera egomotion and produces relative poses. Global trajectories are reconstructed by integrating these relative poses, given an initial state, and thus it is critical to keep motion transformation estimates accurate enough to ensure high-prevision localization in a global scale. This section discusses deep learning approaches to achieve odometry estimation from various sensor data, that are fundamentally different in their data properties and application scenarios.
The discussion mainly focuses on odometry estimation from  visual, inertial and point-cloud data, as they are the common choices of sensing modalities on mobile agents.

\subsection{Visual Odometry}


Visual odometry (VO) estimates the ego-motion of a camera, and integrates the relative motion between images into global poses.
Deep learning methods are capable of extracting high-level feature representations from images, and thereby provide an alternative to solve the VO problem, without requiring hand-crafted feature extractors. 
Existing deep learning based VO models can be categorized into \emph{end-to-end VO} and \emph{hybrid VO}, depending on whether they are purely neural-network based or whether they are a combination of classical VO algorithms and deep neural networks. Depending on the availability of ground-truth labels in the training phase, end-to-end VO systems can be further classified into \emph{supervised} VO and \emph{unsupervised} VO.

\subsubsection{Supervised Learning of VO}

We start with the introduction of supervised VO, one of the most predominant approaches to learning-based odometry, by training a deep neural network model on labelled datasets to construct a mapping function from consecutive images to motion transformations directly, instead of exploiting the geometric structures of images as in conventional VO systems\cite{scaramuzza2011visual}. 
At its most basic, the input of deep neural network is a pair of consecutive images, and the output is the estimated translation and rotation between two frames of images.

One of the first works in this area was Konda et al. \cite{Konda2015}. This approach formulates visual odometry as a classification problem, and predicts the discrete changes of direction and velocity from input images using a convolutional neural network (ConvNet). 
Costante et al. \cite{costante2015exploring} used a ConvNet to extract visual features from dense optical flow, and based on these visual features to output frame-to-frame motion estimation. Nonetheless, these two works have not achieved end-to-end learning from images to motion estimates, and their performance is still limited. 

DeepVO \cite{Wang2017} utilizes a combination of convolutional neural network (ConvNet) and recurrent neural network (RNN) to enable end-to-end learning of visual odometry. The framework of DeepVO becomes a typical choice in realizing supervised learning of VO, due to its specialization in end-to-end learning. 
Figure \ref{fig: vo} (a) shows the architecture of this RNN+ConvNet based VO system, which extracts visual features from pairs of images via a ConvNet, and passes features through RNNs to model the temporal correlation of features. Its ConvNet encoder is based on a FlowNet structure to extract visual features suitable for optical flow and self-motion estimation. 
Using a FlowNet based encoder can be regarded as introducing the prior knowledge of optical flow into the learning process, and potentially prevents DeepVO from being overfitted to the training datasets. The reccurent model summarizes the history information into its hidden states, so that the output is inferred from both past experience and current ConvNet features from sensor observation.
It is trained on large-scale datasets with groundtruthed camera poses as labels. To recover the optimal parameters $\bm{\theta}^{*}$ of framework, the optimization target is to minimize the Mean Square Error (MSE) of the estimated translations $\mathbf{\hat{p}} \in \mathbb{R}^3$ and euler angle based rotations $\hat{\bm{\varphi}} \in \mathbb{R}^3$:
    \begin{equation}
        \label{eq: vo}
        \bm{\theta}^{*} = \argmin_{\bm{\theta}} \frac{1}{N} \displaystyle\sum_{i=1}^{N} \displaystyle\sum_{t=1}^{T} \| \hat{\mathbf{p}}_t - \mathbf{p}_t \|_2^2 + \| \hat{\bm{\varphi}}_t - \bm{\varphi}_t \|_2^2 ,
    \end{equation}
where $(\hat{\mathbf{p}}_t, \hat{\bm{\varphi}}_t)$ are the estimates of relative pose at the timestep $t$, $(\mathbf{p}, \bm{\varphi})$ are the corresponding groundtruth values, $\bm{\theta}$ are the parameters of the DNN framework, $N$ is the number of samples.

DeepVO reports impressive results on estimating the pose of driving vehicles, even in previously unseen scenarios. In the experiment on the KITTI odometry dataset\cite{Geiger2013}, this data-driven solution outperforms conventional representative monocular VO, e.g. VISO2\cite{geiger2011stereoscan} and ORB-SLAM (without loop closure) \cite{Montiel2015}.
Another advantage is that supervised VO naturally produces trajectory with the absolute scale from monocular camera,
while classical VO algorithm is scale-ambiguous using only monocular information. 
This is because deep neural network can implicitly learn and maintain the global scale from large collection of images, which can be viewed as learning from past experience to predict current scale metric.

Based on this typical model of supervised VO, a number of works further extended this approach to improve the model performance. 
To improve the generalization ability of supervised VO, \cite{saputra2019learning} incorporates curriculum learning (i.e. the model is trained by increasing the data complexity) and geometric loss constraints.
Knowledge distillation (i.e. a large model is compressed by teaching a smaller one) is applied into the supervised VO framework to greatly reduce the number of network parameters, making it more amenable for real-time operation on mobile devices \cite{saputra2019distilling}.
Furthermore, Xue et al. \cite{xue2019beyond} introduced a memory module that stores global information, and a refining module that improves pose estimates with the preserved contextual information.

In summary, these end-to-end learning methods benefit from recent advances in machine learning techniques and computational power, to automatically learn pose transformations directly from raw images that can tackle challenging real-world odometry estimation.

\subsubsection{Unsupervised Learning of VO}
\label{sec: unsupervised vo}
There is growing interest in exploring unsupervised learning of VO.
Unsupervised solutions are capable of exploiting unlabelled sensor data, and thus it saves human effort on labelling data, and has better adaptation and generalization ability in new scenarios, where no labelled data are available. 
This has been achieved in a self-supervised framework that jointly learns depth and camera ego-motion from video sequences, by utilizing view synthesis as a supervisory signal\cite{Zhou2017}.

As shown in Figure \ref{fig: vo} (b), a typical unsupervised VO solution consists of a depth network to predict depth maps, and a pose network to produce motion transformations between images. 
The entire framework takes consecutive images as input, and the supervision signal is based on novel view synthesis -
given a source image $\mathbf{I}_s$, the view synthesis task is to generate a synthetic target image $\mathbf{I}_t$. A pixel of source image $\mathbf{I}_s(p_s)$ is projected onto a target view $\mathbf{I}_t(p_t)$ via: 
    \begin{equation}
        \label{eq: un vo}
        p_s \sim \mathbf{K} \mathbf{T}_{t \to s} \mathbf{D}_t(p_t) \mathbf{K}^{-1} p_t
    \end{equation}
where $\mathbf{K}$ is the camera's intrinsic matrix, $\mathbf{T}_{t \to s}$ denotes the camera motion matrix from target frame to source frame, and $\mathbf{D}_t(p_t)$ denotes the per-pixel depth maps in the target frame.
The training objective is to ensure the consistency of the scene geometry by optimizing the photometric reconstruction loss between the real target image and the synthetic one: 
    \begin{equation}
        \mathcal{L}_{\text{photo}} = \sum_{<\mathbf{I}_1, ..., \mathbf{I}_N>\in S} \sum_{p} | \mathbf{I}_t(p) - \hat{\mathbf{I}}_s(p) |,
    \end{equation}
where p denotes pixel coordinates, $\mathbf{I}_t$ is the target image, and $\hat{\mathbf{I}}_s$ is the synthetic target image generated from the source image $\mathbf{I}_s$.

\begin{table*}[t]
\caption{A summary of existing methods on deep learning for odometry estimation.}
\label{tb: odometry estimation}
    \small
    \begin{center}
\begin{tabular}{c l c c c c c l }
\hline
\multicolumn{2}{c}{\multirow{2}{*}{Model}}                   & \multirow{2}{*}{Sensor} & \multirow{2}{*}{Supervision} & \multirow{2}{*}{Scale} & \multicolumn{2}{c}{Performance} & \multirow{2}{*}{Contributions} \\ 
\multicolumn{2}{c}{}                                                       &                         &                              &                        & Seq09                  & Seq10        &          \\ \hline
\multirow{26}{*}{VO}        
& Konda et al.\cite{Konda2015} & MC & Supervised & Yes & - & - & \footnotesize{formulate VO as a classification problem} \\ 
& Costante et al.\cite{costante2015exploring} & MC & Supervised & Yes  & 6.75 & 21.23 & \footnotesize{extract features from optical flow for VO estimates} \\  
& Backprop KF\cite{haarnoja2016backprop} & MC & Hybrid & Yes & - & - & \footnotesize{a differentiable Kalman filter based VO}\\ 
& DeepVO\cite{Wang2017} & MC & Supervised & Yes & - & 8.11 & \footnotesize{combine RNN and ConvNet for end-to-end learning}\\ 
& SfmLearner\cite{Zhou2017} & MC & Unsupervised & No & 17.84 & 37.91 & \footnotesize{novel view synthesis for self-supervised learning} \\
& Yin et al.\cite{yin2017scale} & MC & Hybrid & Yes & 4.14 & 1.70 & \footnotesize{introduce learned depth to recover scale metric} \\  
& UnDeepVO\cite{li2018undeepvo} & SC & Unsupervised & Yes & 7.01 & 10.63 & \footnotesize{use fixed stereo line to recover scale metric} \\ 
& Barnes et al.\cite{barnes2018driven} & MC & Hybrid & Yes & -  & - & \footnotesize{integrate learned depth and ephemeral masks} \\  
& GeoNet\cite{Yin2018} & MC & Unsupervised  & No & 43.76 & 35.6 & \footnotesize{geometric consistency loss and 2D flow generator}  \\ 
& Zhan et al.\cite{Zhan2018} & SC & Unsupervised & No & 11.92  & 12.45 & \footnotesize{use fixed stereo line for scale recovery} \\ 
& DPF\cite{jonschkowski2018differentiable} & MC & Hybrid & Yes & -  & - & \footnotesize{a differentiable particle filter based VO}\\ 
& Yang et al.\cite{yang2018deep} & MC & Hybrid & Yes & 0.83 & 0.74 & \footnotesize{use learned depth into classical VO} \\  
& Zhao et al.\cite{zhao2018learning} & MC & Supervised & Yes & - & 4.38 & \footnotesize{generate dense 3D flow for VO and mapping}  \\ 
& Struct2Depth\cite{casser2019depth} & MC & Unsupervised & No & 10.2 & 28.9 & \footnotesize{introduce 3D geometry structure during learning} \\ 
& Saputra et al.\cite{saputra2019learning} & MC & Supervised & Yes & - & 8.29 & \footnotesize{curriculum learning and geometric loss constraints} \\ 
& GANVO\cite{almalioglu2019ganvo} & MC & Unsupervised & No & - & - & \footnotesize{adversarial learning to generate depth}\\ 
& CNN-SVO\cite{loo2019cnn} & MC & Hybrid & Yes & 10.69 & 4.84 & \footnotesize{use learned depth to initialize SVO}  \\ 
& Xue et al.\cite{xue2019beyond} & MC & Supervised & Yes & - & 3.47 & \footnotesize{memory and refinement module} \\ 
& Wang et al.\cite{wang2019recurrent} & MC & Unsupervised & Yes & 9.30 & 7.21 & \footnotesize{integrate RNN and flow consistency constraint}\\ 
& Li et al.\cite{li2019pose} & MC & Unsupervised & No & - & - & \footnotesize{global optimization for pose graph}                   \\ 
& Saputra et al.\cite{saputra2019distilling} & MC & Supervised & Yes & - & - & \footnotesize{knowledge distilling to compress deep VO model}\\ 
& Gordon\cite{gordon2019depth} & MC & Unsupervised & No & 2.7 & 6.8 & \footnotesize{camera matrix learning}\\ 
& Koumis et al.\cite{koumis2019estimating} & MC & Supervised & Yes  & - & - & \footnotesize{3D convolutional networks}\\ 
& Bian et al.\cite{bian2019unsupervised} & MC & Unsupervised & Yes & 11.2 & 10.1 & \footnotesize{scale recovery from only monocular images}  \\ 
& Zhan et al.\cite{zhan2020visual} & MC & Hybrid & Yes  & 2.61 & 2.29 & \footnotesize{integrate learned optical flow and depth}              \\ 
& D3VO\cite{yang2020d3vo} & MC & Hybrid & Yes & 0.78 & 0.62 & \footnotesize{integrate learned depth, uncertainty and pose} \\ \hline
\multirow{4}{*}{VIO} 
& VINet\cite{Clark2017a} & MC+I & Supervised & Yes & - & - & \footnotesize{formulate VIO as a sequential learning problem}  \\ 
& VIOLearner\cite{shamwell2019unsupervised} & MC+I & Unsupervised & Yes & 1.51 & 2.04 & \footnotesize{online correction module} \\ 
& Chen et al.\cite{chen2019selective} & MC+I & Supervised & Yes & - & - & \footnotesize{feature selection for deep sensor fusion}  \\ 
& DeepVIO\cite{han2019deepvio} & SC+I & Unsupervised & Yes & 0.85 & 1.03 & \footnotesize{learn VIO from stereo images and IMU} \\ \hline
\multirow{4}{*}{LO}           
& Velas et al.\cite{velas2018cnn} & L & Supervised & Yes & 4.94 & 3.27& \footnotesize{ConvNet to estimate odometry from point clouds} \\ 
& LO-Net\cite{li2019net} & L & Supervised & Yes & 1.37 & 1.80 & \footnotesize{geometric constraint loss}                \\ 
& DeepPCO\cite{wang2019deeppco} & L & Supervised & Yes  & - & - & \footnotesize{parallel neural network}             \\ 
& Valente et al.\cite{valente2019deep} & MC+L & Supervised & Yes & - & 7.60 & \footnotesize{sensor fusion for LIDAR and camara} \\ \hline
\end{tabular}
\end{center}
    \begin{itemize}
              \footnotesize{
                \item \textit{Model:} VO, VIO and LO represent visual odometry, visual-inertial odometry and LIDAR odometry respectively.
                \item \textit{Sensor:} MC, SC, I and L represent monocular camera, stereo camera, inertial measurement unit, and LIDAR respectively.
                \item \textit{Supervision} represents whether this work is a purely neural network based model trained with groundtruth labels (Supervised) or without labels (Unsupervised), or it is a combination of classical and deep neural network (Hybrid)
                \item \textit{Scale} indicates whether a trajectory with a global scale can be produced.
                \item \textit{Performance} reports the localization error (a small number is better), i.e. the averaged translational RMSE drift (\%) on lengths of 100m-800m  on the KITTI odometry dataset\cite{Geiger2013}. Most works were evaluated on the Sequence 09 and 10, and thus we took the results on these two sequences from their original papers for a performance comparison. Note that the training sets may be different in each work. 
                \item \textit{Contributions} summarize the main contributions of each work compared with previous research.
            }
    \end{itemize}
\end{table*}

However, there are basically two main problems that remained unsolved in the original work\cite{Zhou2017}: 
1) this monocular image based approach is not able to provide pose estimates in a consistent global scale. Due to the scale ambiguity, no physically meaningful global trajectory can be reconstructed, limiting its real use.
2) The photometric loss assumes that the scene is static and without camera occlusions. Although the authors proposed the use of an explainability mask to remove scene dynamics, the influence of these environmental factors is still not addressed completely, which violates the assumption. 
To address these concerns, an increasing number of works \cite{li2018undeepvo,Yin2018,Zhan2018,yang2018deep,zhao2018learning,almalioglu2019ganvo,li2019pose,li2019sequential,sheng2019unsupervised} extended this unsupervised framework to achieve better performance. 

To solve the global scale problem, \cite{li2018undeepvo,Zhan2018} proposed to utilize stereo image pairs to recover the absolute scale of pose estimation. They introduced an additional spatial photometric loss between the left and right pairs of images, as the stereo baseline (i.e. motion transformation between the left and right images) is fixed and known throughout the dataset. 
Once the training is complete, the network produces pose predictions using only monocular images. Thus, although it is unsupervised in the context of not having access to ground-truth, the training dataset (stereo) is different to the test set (mono). \cite{bian2019unsupervised} tackles the scale issue by introducing a geometric consistency loss, that enforces the consistency between predicted depth maps and reconstructed depth maps. The framework transforms the predicted depth maps into a 3D space, and projects them back to produce reconstructed depth maps. In doing so, the depth predictions are able to remain scale-consistent over consecutive frames, enabling pose estimates to be scale-consistent meanwhile.

The photometric consistency constraint assumes that the entire scenario consists only of rigid static structures, e.g. buildings and lanes. However, in real-world applications, environmental dynamics (e.g. pedestrians and vehicles), will distort the photometric projection and degrade the accuracy of pose estimation. To address this concern, GeoNet \cite{Yin2018} divides its learning process into two sub-tasks by estimating static scene structures and motion dynamics separately through a rigid structure reconstructor and a non-rigid motion localizer. In addition, GeoNet enforces a geometric consistency loss to mitigate the issues caused by camera occlusions and non-Lambertian surfaces. \cite{zhao2018learning} adds a 2D flow generator along with a depth network to generate 3D flow. Benefiting from better 3D understanding of environment, their framework is able to produce more accurate camera pose, along with a point cloud map. 
GANVO \cite{almalioglu2019ganvo} employs a generative adversarial learning paradigm for depth generation, and introduces a temporal recurrent module for pose regression.
Li et al. \cite{li2019sequential} also utilized a generative adversarial network (GAN) to generate more realistic depth maps and poses, and further encourage more accurate synthetic images in the target frame. Instead of a hand-crafted metric, a discriminator is adopted to evaluate the quality of synthetic images generation. In doing so, the generative adversarial setup facilitates the generated depth maps to be more texture-rich and crisper. In this way, high-level scene perception and representation are accurately captured and environmental dynamics are implicitly tolerated. 

Although unsupervised VO still cannot compete with supervised VO in performance, as illustrated in Figure \ref{fig: odometry performance}, its concerns of scale metric and scene dynamics problem have been largely resolved. With the benefits of self-supervised learning, and ever-increasing improvement on performance, unsupervised VO would be a promising solution in providing pose information, and tightly coupled with other modules in spatial machine intelligence system.

\subsubsection{Hybrid VO}
Unlike end-to-end VO that only relies on a deep neural network to interpret pose from data, hybrid VO integrates classical geometric models with deep learning framework. Based on mature geometric theory, they use a deep neural network to expressively replace parts of a geometry model.

A straightforward way is to incorporate the learned depth estimates into a conventional visual odometry algorithm to recover the absolute scale metric of poses \cite{yin2017scale}. Learning depth estimation is a well-researched area in the computer vision community. For example, \cite{eigen2014depth,ummenhofer2017demon,garg2016unsupervised,godard2017unsupervised} provide per-pixel depths in a global scale by employing a trained deep neural model. Thus the so-called scale problem of conventional VO is mitigated. 
Barnes et al. \cite{barnes2018driven} utilize both the predicted depth maps and ephemeral masks (i.e. the area of moving objects) into a VO system to improve its robustness to moving objects.  
Zhan et al. \cite{zhan2020visual} integrate the learned depth and optical flow predictions into a conventional visual odometry model, achieving competitive performance over other baselines.
Other works combine physical motion models with deep neural network e.g. via a differentiable Kalman filter \cite{Haarnoja2016}, and a particle filter \cite{Jonschkowski2018}. The physical model serves as an algorithmic prior in the learning process.
Furthermore, D3VO \cite{yang2020d3vo} incorporates the deep predictions of depth, pose, and uncertainty into a direct visual odometry. 

Combining the benefits from both geometric theory and deep learning, hybrid models are normally more accurate than end-to-end VO at this stage, as shown in Table 1.
It is notable that hybrid models even outperform the state-of-the-art conventional monocular VO or visual-inertial odometry (VIO) systems on common benchmarks, for example, D3VO\cite{yang2020d3vo} defeats several popular conventional VO/VIO systems, such as DSO\cite{vi-dso}, ORB-SLAM\cite{Montiel2015}, VINS-Mono\cite{Qin2018}. This demonstrates the rapid rate of progress in this area.

\subsection{Visual-Inertial Odometry}

Integrating visual and inertial data as visual-inertial odometry (VIO) is a well-defined problem in mobile robotics.
Both cameras and inertial sensors are relatively low-cost, power-efficient and widely deployed. These two sensors are complementary: monocular cameras capture the appearance and structure of a 3D scene, while they are scale-ambiguous, and not robust to challenging scenarios, e.g. strong lighting changes, lack of texture and high-speed motion; 
In contrast, IMUs are completely ego-centric, scene-independent, and can also provide absolute metric scale. Nevertheless, the downside is that inertial measurements, especially from low-cost devices, are plagued by process noise and biases. 
An effective fusion of the measurements from these two complementary sensors is of key importance to accurate pose estimation. Thus, according to their information fusion methods, conventional model based visual-inertial approaches are roughly segmented into three different classes :
filtering approaches \cite{Li2013b}, fixed-lag smoothers \cite{Leutenegger2015} and full smoothing methods \cite{Forster2017}.

Data-driven approaches have emerged to consider learning 6-DoF poses directly from visual and inertial measurements without human intervention or calibration.
VINet \cite{Clark2017a} is the first work that formulated visual-inertial odometry as a sequential learning problem, and proposed a deep neural network framework to achieve VIO in an end-to-end manner.
VINet uses a ConvNet based visual encoder to extract visual features from two consecutive RGB images, and an inertial encoder to extract inertial features from a sequence of IMU data with a long short-term memory (LSTM) network. 
Here, the LSTM aims to model the temporal state evolution of inertial data. 
The visual and inertial features are concatenated together, and taken as the input into a further LSTM module to predict relative poses, conditioned on the history of system states.
This learning approach has the advantage of being more robust to calibration and relative timing offset errors. 
However, VINet has not fully addressed the problem of learning a meaningful sensor fusion strategy. 

To tackle the deep sensor fusion problem, Chen et al. \cite{chen2019selective} proposed selective sensor fusion, a framework that selectively learns context-dependent representations for visual inertial pose estimation. 
Their intuition is that the importance of features from different modalities should be considered according to the exterior (i.e., environmental) and interior (i.e., device/sensor) dynamics, by fully exploiting the complementary behaviors of two sensors.
Their approach outperforms those without a fusion strategy, e.g. VINet, avoiding catastrophic failures.

Similar to unsupervised VO,
Visual-inertial odometry can also be solved in a self-supervised fashion using  novel view synthesis. 
VIOLearner \cite{shamwell2019unsupervised} constructs motion transformations from raw inertial data, and converts source images into target images with the camera matrix and depth maps via the Equation \ref{eq: un vo} mentioned in Section \ref{sec: unsupervised vo}. 
In addition, an online error correction module corrects the intermediate errors of the framework. 
The network parameters are recovered by optimizing a photometric loss.
Similarly, DeepVIO \cite{han2019deepvio} incorporates inertial data and stereo images into this unsupervised learning framework, and is trained with a dedicated loss to reconstruct trajectories in a global scale. 

Learning-based VIO cannot defeat the state-of-the-art classical model based VIOs, but they are generally more robust to real issues\cite{Clark2017a,chen2019selective,han2019deepvio} such as measurement noises, bad time synchronization, thanks to the impressive ability of DNNs in feature extraction and motion modelling. 

\subsection{Inertial Odometry}
Beyond visual odometry and visual-inertial odometry, an inertial-only solution, i.e. inertial odometry provides an ubiquitous alternative to solve the odometry estimation problem. Compared with visual methods, an inertial sensor is relatively low-cost, small, energy efficient and privacy preserving. It is relatively immune to environmental factors, such as lighting conditions or moving objects. However, low-cost MEMS inertial measurement units (IMU) widely found on robots and mobile devices are corrupted with high sensor bias and noise, leading to unbounded error drifts in the strapdown inertial navigation system (SINS), if inertial data are doubly integrated.

Chen et al. \cite{chen2018ionet} formulated inertial odometry as a sequential learning problem with a key observation that 2D motion displacements in the polar coordinate (i.e. polar vector) can be learned from independent windows of segmented inertial data. The key observation is that when tracking human and wheeled configurations, the frequency of their vibrations is relevant to the moving speed, which is reflected by inertial measurements. 
Based on this, they proposed IONet, a LSTM based framework for end-to-end learning of relative poses from sequences of inertial measurements. Trajectories are generated by integrating motion displacements. 
\cite{chen2019motiontransformer} leveraged deep generative models and domain adaptation technique to improve the generalization ability of deep inertial odometry in new domains. 
\cite{esfahani2019aboldeepio} extends this framework by an improved triple-channel LSTM network to predict polar vectors for drone localization from inertial data and sampling time.
RIDI \cite{yan2018ridi} trains a deep neural network to regress linear velocities from inertial data, calibrates the collected accelerations to satisfy the constraints of the learned velocities, and doubly integrates the accelerations into locations with a conventional physical model.
Similarly, \cite{cortes2018deep} compensates the error drifts of the classical SINS model with the aid of learned velocities. 
Other works have also explored the usage of deep learning to detect zero-velocity phase for navigating pedestrians \cite{wagstaff2018lstm} and vehicles \cite{brossard2019rins}. This zero-velocity phase provides context information to correct system error drifts via Kalman filtering. 

Inertial only solution can be a backup plan to offer pose information in extreme environments, where visual information is not available or is highly distorted. Deep learning has proven its capability to learn useful features from noisy IMU data, and compensate the error drifts of inertial dead reckoning, which is difficult to solve by classical algorithms.  

\begin{figure}
    \centering
    \includegraphics[width=0.5\textwidth]{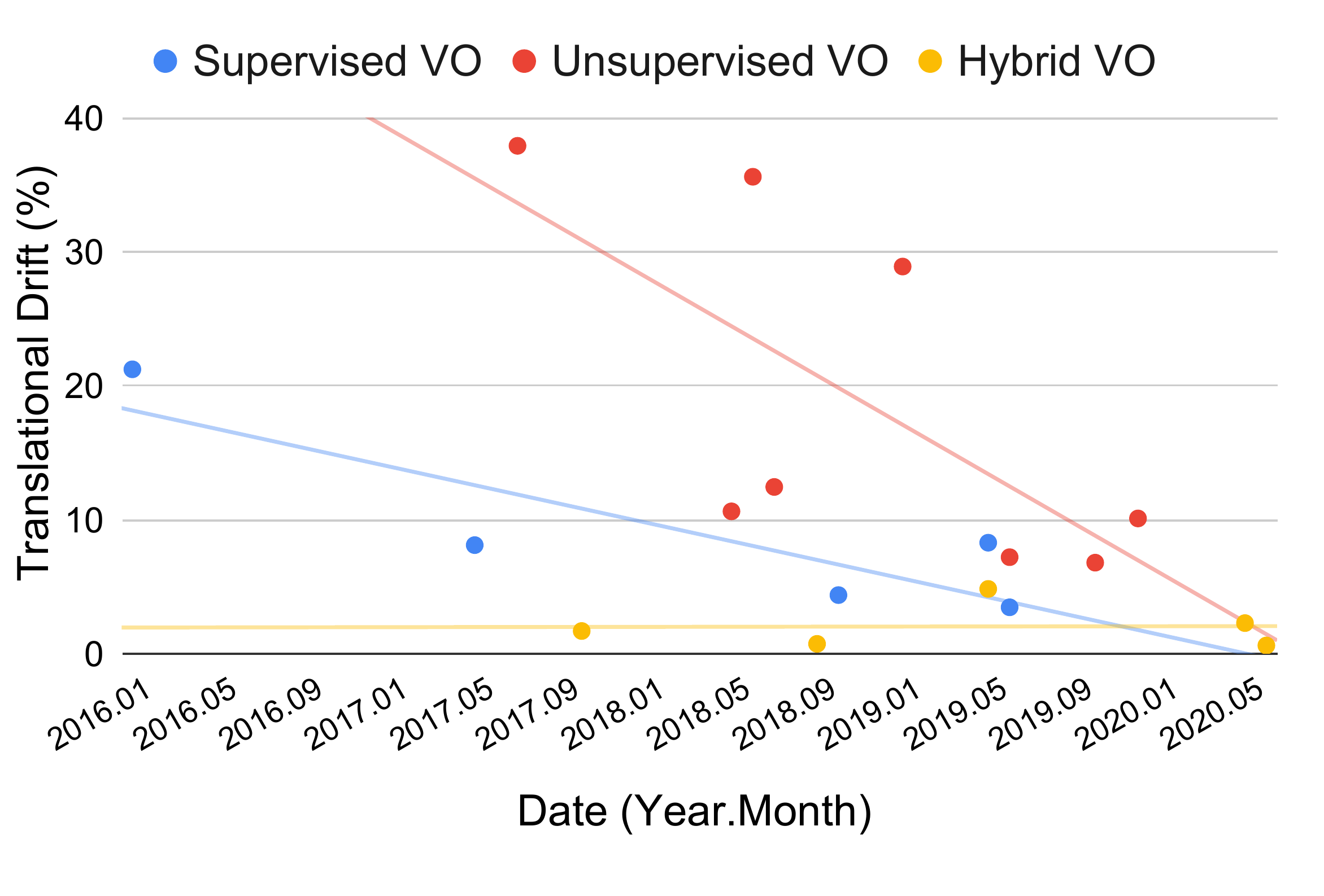}
    \caption{A comparison of the performance of deep learning based visual odometry with an evaluation on the Trajectory 10 of the KITTI dataset. 
    }
    \label{fig: odometry performance}
\end{figure}

\subsection{LIDAR Odometry}

LIDAR sensors provide high-frequency range measurements, with the benefits of working consistently in complex lighting conditions and optically featureless scenarios. Mobile robots and self-driving vehicles are normally equipped with LIDAR sensors to obtain relative self-motion (i.e. LIDAR odometry) and global pose with respect to a 3D map (LIDAR relocalization). The performance of LIDAR odometry is sensitive to point cloud registration errors due to non-smooth motion. In addition, the data quality of LIDAR measurements is also affected by extreme weather conditions, for example, heavy rain or fog/mist.

Traditionally, LIDAR odometry 
 relies on  point cloud registration to detect feature points, e.g. line and surface segments, and uses a matching algorithm to obtain the pose transformation by minimizing the distance between two consecutive point-cloud scans. 
Data-driven methods consider solving LIDAR odometry in an end-to-end fashion, by leveraging deep neural networks to construct a mapping function from point cloud scan sequences to pose estimates \cite{velas2018cnn,li2019net,wang2019deeppco}.
As point cloud data are challenging to be directly ingested by neural networks due to their sparse and irregularly sampled format,
these methods typically convert point clouds into a regular matrix through cylindrical projection, and adopt ConvNets to extract features from consecutive point cloud scans. These networks regress relative poses and are trained via ground-truth labels.
LO-Net \cite{li2019net} reports competitive performance over the conventional state-of-the-art algorithm, i.e. the LIDAR Odometry and Mapping (LOAM) algorithm \cite{Zhang2010}. 




\subsection{Comparison of Odometry Estimation}

Table \ref{tb: odometry estimation} compares existing work on odometry estimation, in terms of their sensor type, model, whether a trajectory with an absolute scale is produced, and their performance evaluation on the KITTI dataset, where available. As deep inertial odometry has not been evaluated on the KITTI dataset, we do not include inertial odometry in this table. The KITTI dataset \cite{Geiger2013} is a common benchmark for odometry estimation, consisting of a collection of sensor data from car-driving scenarios. As most data-driven approaches adopt the trajectory 09 and 10 of the KITTI dataset to evaluate model performance, we compared them according to the averaged Root Mean Square Error (RMSE) of the translation for all the subsequences of lengths (100, 200, .., 800) meters, which is provided by the official KITTI VO/SLAM evaluation metrics.

We take visual odometry as an example. Figure \ref{fig: odometry performance} reports the translational drifts of deep visual odometry models over time on the 10th trajectory of the KITTI dataset. Clearly, hybrid VO shows the best performance over supervised VO and unsupervised VO, as the hybrid model benefits from both the mature geometry models of traditional VO algorithms and the strong capacity for feature extraction of deep learning. Although supervised VO still outperforms unsupervised VO, the performance gap between them is diminishing as the limitations of unsupervised VO are gradually addressed. For example, it has been found that unsupervised VO now can recover global scale from monocular images \cite{bian2019unsupervised}. Overall, data-driven visual odometry shows a remarkable increase in model performance, indicating the potentials of deep learning approaches in achieving more accurate odometry estimation in the future.


\section{Mapping}

Mapping refers to the ability of a mobile agent to build a consistent environmental model to describe the surrounding scene. 
Deep learning has fostered a set of tools for scene perception and understanding, with applications ranging from depth prediction, to semantic labelling, to 3D geometry reconstruction.  
This section provides an overview of existing works relevant to deep learning based mapping methods. We categorize them into geometric mapping, semantic mapping, and general mapping. Table \ref{tb: mapping} summarizes the existing methods on deep learning based mapping.

\begin{table*}[t]
\caption{A summary of existing methods on deep learning for mapping.}
\label{tb: mapping}
    \small
\begin{tabular}{c c l}
 \hline
                               & Output Representation & Employed by \\
                               \hline
\multirow{4}{*}{Geometric Map} & Depth Representation           &  \cite{eigen2014depth,liu2015learning,ummenhofer2017demon}, \cite{garg2016unsupervised,godard2017unsupervised}, \cite{Zhou2017}, \cite{li2018undeepvo,Yin2018,Zhan2018,yang2018deep,zhao2018learning,wang2018learning,almalioglu2019ganvo,li2019pose}, \cite{li2019sequential,sheng2019unsupervised}      \\
                               & Voxel Representation          &  \cite{ji2017surfacenet,paschalidou2018raynet}, \cite{kar2017learning}(Object), \cite{tatarchenko2017octree}, \cite{hane2017hierarchical}(Object), \cite{dai2017shape}(Object), \cite{riegler2017octnetfusion}(Object)
           \\
                               & Point Representation          &  \cite{fan2017point}(Object)           \\
                               & Mesh Representation          &   \cite{groueix2018papier}(Object), \cite{wang2018pixel2mesh}(Object), \cite{ladicky2017point}(Object), \cite{dai2019scan2mesh}(Object), \cite{mukasa20173d,bloesch2019learning}         \\
                               \hline
\multirow{3}{*}{Semantic Map}  & Semantic Segmentation   &  \cite{mccormac2017semanticfusion,ma2017multi,xiang2017rnn}           \\
                               & Instance Segmentation   &  \cite{sunderhauf2017meaningful,mccormac2018fusion++,grinvald2019volumetric}           \\
                               & Panoptic Segmentation   &  \cite{narita2019panopticfusion}          \\
                               \hline
General Map                   & Neural Representation     &  \cite{Bloesch2018},\cite{tobin2019geometry,lim2019neural,sitzmann2019scene}, \cite{jaderberg2017reinforcement,mirowski2016learning,zhu2017target,mirowski2018learning}          \\
                               \hline
\end{tabular}
\begin{itemize}
              \footnotesize{
                \item \textit{Object} indicates that this method is only validated on reconstructing single objects rather than a scene. 
            }
    \end{itemize}
\end{table*}



\subsection{Geometric Mapping}
\begin{figure}
    \centering
    \includegraphics[width=0.45\textwidth]{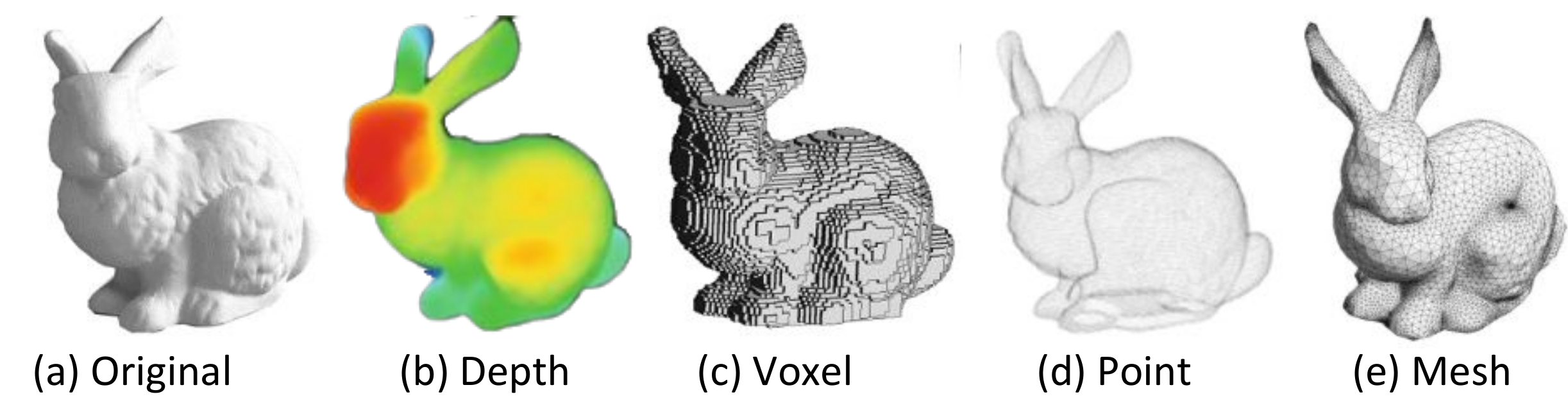}
    \caption{An illustrations of scene representations on the Stanford Bunny benchmark: (a) original model, (b) depth representation, (c) voxel representation (d) point representation (e) mesh representation.}
    \label{fig: scene represenation}
\end{figure}

Broadly, geometric mapping captures the shape and structural description of a scene. 
Typical choices of the scene representations used in geometric mapping include depth, voxel, point and mesh. We follow this representational taxonomy and categorize deep learning for geometric mapping into the above four classes. Figure \ref{fig: scene represenation} demonstrates these geometric representations on the Stanford Bunny benchmark.


\subsubsection{Depth Representation}

Depth maps play a pivotal role in understanding the scene geometry and structure. Dense scene reconstruction has been achieved by fusing depth and RGB images \cite{kerl2013dense,whelan2015real}. Traditional SLAM systems represent scene geometry with dense depth maps (i.e. 2.5D), such as DTAM \cite{Newcombe2011}. In addition, accurate depth estimation can contribute to the absolute scale recovery for visual SLAM. 

Learning depth from raw images is a fast evolving area in computer vision community. 
The earliest work formulates depth estimation as a mapping function of input single images, constructed by a multi-scale deep neural network \cite{eigen2014depth} to output the per-pixel depth maps from single images. More accurate depth prediction is achieved by jointly optimizing the depth and self-motion estimation \cite{ummenhofer2017demon}. These supervised learning methods \cite{eigen2014depth,liu2015learning,ummenhofer2017demon} can predict per-pixel depth by training deep neural networks on large data collections of images with corresponding depth labels.   
Although they are found outperforming the traditional structure based methods, such as \cite{karsch2014depth}, their effectiveness are largely reliant on model training and can be difficult to generalize to new scenarios in absence of labeled data.

On the other side, recent advances in this field focus on unsupervised solutions, by reformulating depth prediction as a novel view synthesis problem.
\cite{garg2016unsupervised,godard2017unsupervised} utilized photometric consistency loss as a self-supervision signal for training neural models. With stereo images and a known camera baseline, \cite{garg2016unsupervised,godard2017unsupervised} synthesize the left view from the right image, and the predicted depth maps of the left view. By minimizing the distance between synthesized images and real images, i.e. the spatial consistency, the parameters of the networks can be recovered via this self-supervision in an end-to-end manner. 
Besides the spatial consistency, \cite{Zhou2017} proposed to apply temporal consistency as a self-supervised signal, by synthesizing the image in the target time frame from the source time frame. At the same time, egomotion is recovered along with the depth estimation. This framework only requires monocular images to learn both the depth maps and egomotion. A number of following works \cite{li2018undeepvo,Yin2018,Zhan2018,yang2018deep,zhao2018learning,wang2018learning,almalioglu2019ganvo,li2019pose,li2019sequential,sheng2019unsupervised} extended this framework and achieved better performance in depth and egomotion estimation. We refer the readers to Section \ref{sec: unsupervised vo}, in which a variety of additional constraints haven been discussed.


With the depth maps predicted by ConvNets, learning based SLAM systems can integrate depth information to address some limitations of classical monocular solution. 
For example, CNN-SLAM \cite{tateno2017cnn} utilizes the learned depths from single images into a monocular SLAM framework (i.e. LSD-SLAM \cite{engel2014lsd}). Their experiment shows how the learned depth maps contribute to mitigate the absolute scale recovery problem in pose estimates and scene reconstruction. CNN-SLAM achieves dense scene predictions even in texture-less areas, which is normally hard for a conventional SLAM system.


\subsubsection{Voxel Representation}
Voxel-based formulation is a natural way to represent 3D geometry. Similar to the usage of pixel (i.e. 2D element) in images, voxel is a volume element in a three-dimensional space. 
Previous works have explored to use multiple input views, to reconstruct the volumetric representation of scene \cite{ji2017surfacenet,paschalidou2018raynet} and objects \cite{kar2017learning}. For example, SurfaceNet \cite{ji2017surfacenet} learns to predict the confidence of a voxel to determine whether it is on surface or not, and reconstruct the 2D surface of a scene. RayNet \cite{paschalidou2018raynet} reconstructs the scene geometry by extracting view-invariant features while imposing geometric constraints. 
Recent works focus on generating high-resolution 3D volumetric models \cite{hane2017hierarchical,tatarchenko2017octree}. For example, Tatarchenko et al. \cite{tatarchenko2017octree} designed a convolutional decoder based on octree-based formulation to enable scene reconstruction in much higher resolution. Other work can be found on scene completion from RGB-D data \cite{dai2017shape,riegler2017octnetfusion}.
One limitation of voxel representation is the high computational requirement, especially when attempting to reconstruct a scene in high resolution.

\subsubsection{Point Representation}
Point-based formulation consists of the 3-dimensional coordinates (x, y, z) of points in 3D space.
Point representation is easy to understand and manipulate, but suffers from the ambiguity problem, which means that different forms of point clouds can represent a same geometry.
The pioneer work, PointNet \cite{qi2017pointnet}, processes unordered point data with a single symmetric function - max pooling, to aggregate point features for classification and segmentation.
Fan et al. \cite{fan2017point} developed a deep generative model that generates 3D geometry in point-based formulation from single images. In their work, a loss function based on Earth Mover's distance is introduced to tackle the problem of data ambiguity. However, their method is only validated on the reconstruction task of single objects. No work on point generation for scene reconstruction has been found yet.

\subsubsection{Mesh Representation}

Mesh-based formulation encodes the underlying structure of 3D models, such as edges, vertices and faces. It is a powerful representation that naturally captures the surface of 3D shape.
Several works considered the problem of learning mesh generation from images \cite{groueix2018papier,wang2018pixel2mesh} or point clouds data \cite{ladicky2017point,dai2019scan2mesh}. However, these approaches are only able to reconstruct single objects, and limited to generating models with simple structures or from familiar classes. 
To tackle the problem of scene reconstruction in mesh representation, \cite{mukasa20173d} integrates the sparse features from monocular SLAM with dense depth maps from ConvNet for the update of 3D mesh representation. The depth predictions are fused into the monocular SLAM system to recover the absolute scale of pose and scene features estimation. 
To allow efficient computation and flexible information fusion, \cite{bloesch2019learning} utilizes 2.5D mesh to represent scene geometry. In their approach, the image plane coordinates of mesh vertices are learned by deep neural networks, while the depth maps are optimized as free variables. 


\subsection{Semantic Map}



\begin{figure}
    \centering
    \includegraphics[width=0.45\textwidth]{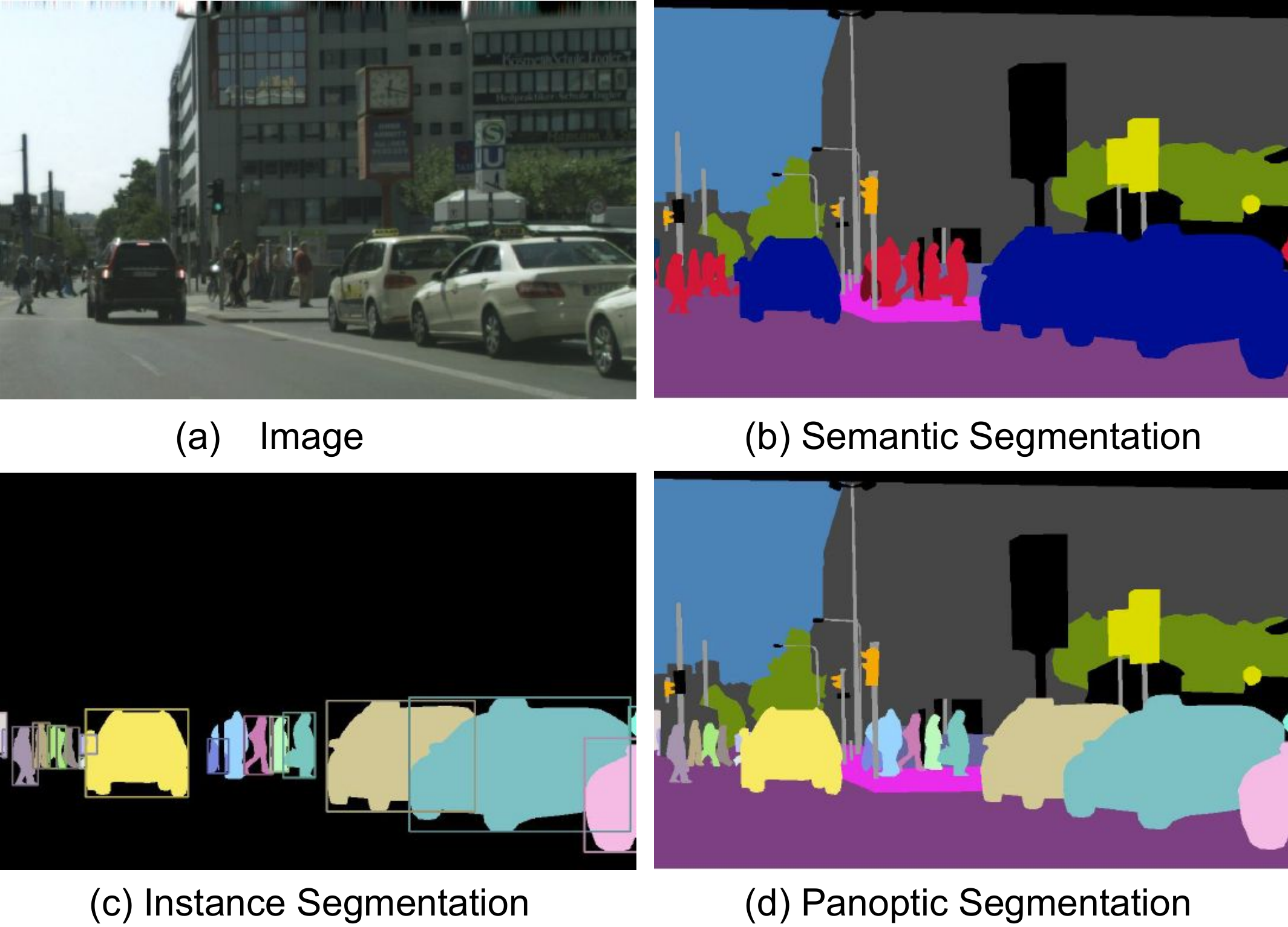}
    \caption{(b) semantic segmentation, (c) instance segmentation and (d) panoptic segmentation for semantic mapping \cite{kirillov2019panoptic}.}
    \label{fig: semantic mapping}
\end{figure}

Semantic mapping connects semantic concepts (i.e. object classification, material composition etc) with the geometry of environments. This is treated as a data association problem.
The advances in deep learning greatly fosters the developments of object recognition and semantic segmentation. 
Maps with semantic meanings enable mobile agents to have high-level understandings of their environments beyond pure geometry, and allow for a greater range of functionality and autonomy. 

SemanticFusion \cite{mccormac2017semanticfusion} is one of the early works that combined the semantic segmentation labels from deep ConvNet with the dense scene geometry from a SLAM system. It incrementally integrates per-frame semantic segmentation predictions into a dense 3D map by probabilistically associating the 2D frames with the 3D map. This combination not only generates a map with useful semantic information, but also shows the integration with a SLAM system helps to enhance the single frame segmentation. The two modules are loosely coupled in SemanticFusion.
\cite{ma2017multi} proposed a self-supervised network that predicts consistent semantic labels for a map, by imposing constraints on the consistency of semantic predictions in multiple views.
DA-RNN \cite{xiang2017rnn} introduces recurrent models into semantic segmentation framework to learn the temporal connections over multiple view frames, producing more accurate and consistent semantic labelling for volumetric maps from KinectFusion \cite{newcombe2011kinectfusion}.
Yet these methods provide no information on object instances, which means that they are not able to distinguish among different ojects from the same category.

With the advances in instance segmentation, semantic mapping evolves into the instance level. A good example is
\cite{sunderhauf2017meaningful} that offers object-level semantic mapping by identifying individual objects via a bounding box detection module and an unsupervised geometric segmentation module.
Unlike other dense semantic mapping approaches, Fusion++ \cite{mccormac2018fusion++} builds a semantic graph-based map, which predicts only object instances and maintains a consistent map via loop closure detection, pose-graph optimization and further refinement. 
\cite{grinvald2019volumetric} presented a framework that achieves instance-aware semantic mapping, and enables novel object discovery.
Recently, panoptic segmentation \cite{kirillov2019panoptic} attracts a lot of attentions. PanopticFusion \cite{narita2019panopticfusion} advanced semantic mapping to the level of stuff and things level that classifies static objects, e.g. walls, doors, lanes as stuff classes, and other accountable objects as things classes, e.g. moving vehicles, human and tables. Figure \ref{fig: semantic mapping} compares semantic segmentation, instance segmentation and panoptic segmentation.


\subsection{General Map}


Beyond the explicit geometric and semantic map representation, deep learning models are able to encode the whole scene into an \textit{implicit} representation, i.e. a general map representation to capture the underlying scene geometry and appearance.

Utilizing deep autoencoders can automatically discover the high-level compact representation of high-dimensional data. A notable example is CodeSLAM \cite{Bloesch2018} that encodes observed images into a compact and optimizable representation to contain the essential information of a dense scene. This general representation is further used into a keyframe-based SLAM system to infer both pose estimates and keyframe depth maps. Due to the reduced size of learned representations, CodeSLAM allows efficient optimization of tracking camera motion and scene geometry for a global consistency.

Neural rendering models are another family of works that learn to model 3D scene structure implicitly by exploiting view synthesis as a self-supervision signal. 
The target of neural rendering task is to reconstruct a new scene from an unknown viewpoint.
The seminar work, Generative Query Network (GQN) \cite{eslami2018neural} learns to capture representation and render a new scene. 
GQN consists of a representation network and a generation network: the representation network encodes the observations from reference views into a scene representation; the generation network which is based on recurrent model, reconstructs the scene from a new view conditioned on the scene representation and a stochastic latent variable. Taking inputs as the observed images from several viewpoints, and the camera pose of a new view, GQN predicts the physical scene of this new view. Intuitively, through end-to-end training, the representation network can capture the necessary and important factors of 3D environment for the scene reconstruction task via the generation network. GQN is extended by incorporating a geometric-aware attention mechanism to allow more complex environment modelling \cite{tobin2019geometry}, as well as including multimodal data for scene inference \cite{lim2019neural}. Scene representation network (SRN) \cite{sitzmann2019scene} tackles the scene rendering problem via a learned continuous scene representation that connects a camera pose and its corresponding observation. A differentiable Ray Marching algorithm is integrated into SRN to enforce the network to model 3D structure consistently. 
However, these frameworks can only be applied to synthetic datasets due to the complexity of real-world environments.

Last but not least, in the quest of `map-less' navigation, task-driven maps emerge as a novel map representation. This representation is jointly modelled by deep neural networks with respect to the task at hand. Generally those tasks leverage location information, such as navigation or path planning, requiring mobile agents to understand the geometry and semantics of environment. 
Navigation in unstructured environments (even in a city scale) is formulated as an policy learning problem in these works  \cite{jaderberg2017reinforcement,mirowski2016learning,zhu2017target,mirowski2018learning}, and solved by deep reinforcement learning. 
Different from traditional solutions that follow a procedure of building an explicit map, planning path and making decisions, these learning based techniques predict control signals directly from sensor observations in an end-to-end manner, without explicitly modelling the environment. The model parameters are optimized via sparse reward signals, for example, whenever agents reach a destination, a positive reward will be given to tune the neural network.
Once a model is trained, the actions of agents can be determined conditioned on the current observations of environment, i.e. images. In this case, all of environmental factors, such as the geometry, appearance and semantics of a scene, are embedded inside the neurons of a deep neural network and suitable for solving the task at hand.
Interestingly, the visualization of the neurons inside a neural model that is trained on the navigation task via reinforcement learning, has similar patterns as the grid and place cells inside human brain. This provides cognitive cues to support the effectiveness of neural map representation. 




\section{Global Localization}

Global localization concerns the retrieval of absolute pose of a mobile agent within a known scene. 
Different from odometry estimation that relies on estimating the internal dynamical model and can perform in an unseen scenario, in global localization, prior knowledge about the scene is provided and exploited, through a 2D or 3D scene model.
Broadly, it describes the relation between the sensor observations and map, by matching a query image or view against a pre-built model, and returning an estimate of global pose. 

We categorize deep learning based global localization into three categories, according to the types of inquiry data and map: \emph{2D-to-2D localization} queries 2D images against an explicit database of geo-referenced images or implicit neural map; \emph{2D-to-3D localization} establishes correspondences between 2D pixels of images and 3D points of a scene model; and \emph{3D-to-3D localization} matches 3D scans to a pre-built 3D map. 
Table \ref{tb: learning for 2d-2d localization}, \ref{tb: learning for 2d-3d localization} and \ref{tab: 3d-3d localization} summarize the existing approaches on deep learning based 2D-to-2D localization, 2D-to-3D localization and 3D-to-3D localization respectively. 

\subsection{2D-to-2D Localization}
2D-to-2D localization regresses the camera pose of an image against a 2D map. Such 2D map is explicitly built by a geo-referenced database or implicitly encoded in a neural network.

\begin{figure}
    \centering
    \includegraphics[width=0.4\textwidth]{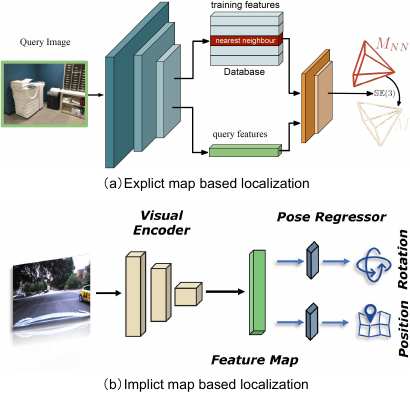}
    \caption{The typical architectures of 2D-to-2D based localization through (a) explict map, i.e. RelocNet \cite{balntas2018relocnet} and (b) implicit map, i.e. e.g. PoseNet \cite{kendall2015posenet}}
    \label{fig: 2d2d}
\end{figure}


\begin{table*}[t]
\caption{A summary on existing methods on deep learning for 2D-to-2D global localization}
\label{tb: learning for 2d-2d localization}
    \small
    \begin{center}
\begin{tabular}{c c l c c c l }
\hline
\multicolumn{3}{c}{\multirow{2}{*}{Model}} & \multirow{2}{*}{Agnostic} & \multicolumn{2}{c}{Performance (m/degree)} & \multicolumn{1}{c}{\multirow{2}{*}{Contributions}} \\
\multicolumn{3}{c}{} & & 7Scenes & Cambridge & \multicolumn{1}{c}{}\\ \hline
\multirow{22}{*}{\rotatebox{90}{2D-to-2D Localization}}
& \multirow{5}{*}{\rotatebox{90}{Explicit Map}} 
& NN-Net  \cite{laskar2017camera} & Yes & 0.21/9.30  & - & combine retrieval and relative pose estimation \\
&& DeLS-3D \cite{wang2018dels} & No  & -  & - & jointly learn with semantics \\ 
&& AnchorNet \cite{saha2018improved} & Yes & 0.09/6.74  & 0.84/2.10 & anchor point allocation \\
&& RelocNet \cite{balntas2018relocnet} & Yes & 0.21/6.73 & - & camera frustum overlap loss \\
&& CamNet \cite{ding2019camnet} & Yes & 0.04/1.69  & - & multi-stage image retrieval \\ \cline{2-7}
& \multirow{17}{*}{\rotatebox{90}{Implicit Map}}                                
& PoseNet \cite{kendall2015posenet} & No  & 0.44/10.44 & 2.09/6.84 & first neural network in global pose regression \\ 
&& Bayesian PoseNet \cite{kendall2016modelling} & No  & 0.47/9.81  & 1.92/6.28 & estimate Bayesian uncertainty for global pose \\ 
&& BranchNet \cite{wu2017delving} & No  & 0.29/8.30  & - & multi-task learning for orientation and translation\\ 
&& VidLoc \cite{Clark2017} & No  & 0.25/- & - & efficient localization from image sequences\\ 
&& Geometric PoseNet \cite{kendall2017geometric} & No  & 0.23/8.12  & 1.63/2.86 & geometry-aware loss \\
&& SVS-Pose \cite{naseer2017deep} & No  & - & 1.33/5.17 & data augmentation in 3D space \\  
&& LSTM PoseNet \cite{walch2017image} & No  & 0.31/9.85  & 1.30/5.52 & spatial correlation \\ 
&& Hourglass PoseNet \cite{melekhov2017image} & No  & 0.23/9.53  & - & hourglass-shaped architecture \\ 
&& VLocNet \cite{valada2018deep} & No  & 0.05/3.80  & 0.78/2.82 & jointly learn global localization and odometry \\ 
&& MapNet \cite{Brahmbhatt2018} & No  & 0.21/7.77  & 1.63/3.64 & impose spatial and temporal constraints\\ 
&& SPP-Net \cite{purkait2018synthetic} & No  & 0.18/6.20  & 1.24/2.68 & synthetic data augmentation \\ 
&& GPoseNet \cite{cai2018hybrid} & No  & 0.30/9.90  & 2.00/4.60 & hybrid model with Gaussian Process Regressor \\
&& VLocNet++ \cite{radwan2018vlocnet++} & No  & 0.02/1.39  & - & jointly learn with odometry and semantics \\ 
&& LSG \cite{xue2019local} & No  & 0.19/7.47  & -  & odometry-aided localization\\  
&& PVL \cite{huang2019prior} & No & - & 1.60/4.21 & prior-guided dropout mask to improve robustness\\ 
&& AdPR \cite{bui2019adversarial} & No  & 0.22/8.8   & -  & adversarial architecture\\ 
&& AtLoc \cite{wang2019atloc} & No  & 0.20/7.56  & -  & attention-guided spatial correlation            
\\ \hline
\end{tabular}
\end{center}
\begin{itemize}
              \footnotesize{
                \item \textit{Agnostic} indicates whether it can generalize to new scenarios.
                \item \textit{Performance} reports the position (m) and orientation (degree) error (a small number is better) on the 7-Scenes (Indoor)\cite{shotton2013scene} and Cambridge (Outdoor) dataset\cite{kendall2015posenet}. Both datasets are split into training and testing set. We report the averaged error on the testing set.
                \item \textit{Contributions} summarize the main contributions of each work compared with previous research.
            }
    \end{itemize}
\end{table*}

\subsubsection{Explicit Map Based Localization}

Explicit map based 2D-to-2D localization typically represents the scene by a database of geo-tagged images (references) \cite{arandjelovic2014dislocation,chen2011city,torii201524}. Figure \ref{fig: 2d2d} (a) illustrates the two stages of this localization with 2D references: 
image retrieval determines the most relevant part of a scene represented by reference images to the visual queries; pose regression obtains the relative pose of query image with respect to the reference images. 

One problem here is how to find suitable image descriptors for image retrieval. 
Deep learning based approaches \cite{chen2014convolutional,sunderhauf2015performance} are based on a pre-trained ConvNet model to extract image-level features, and then use these features to evaluate the similarities against other images. 
In challenging situations, local descriptors are first extracted, followed by being aggregated to obtain robust global descriptors. 
A good example is NetVLAD \cite{arandjelovic2016netvlad} that designs a trainable generalized VLAD (the Vector of Locally
Aggregated Descriptors) layer. This VLAD layer can be plugged into the off-the-shelf ConvNet architecture to encourage better descriptors learning for image retrieval. 

In order to obtain more precise poses of the queries, additional relative pose estimation with respect to the retrieved images is required. Traditionally, relative pose estimation is tackled by epipolar geometry, relying on the 2D-2D correspondences determined by local descriptors \cite{zhou2019learn, melekhov2019dgc}. In contrast, deep learning approaches regress the relative poses straightforwardly from pairwise images. For example, NN-Net \cite{laskar2017camera} utilized neural network to estimate the pairwise relative poses between the query and the top N ranked references. A triangulation-based fusion algorithm coalesces the predicted N relative poses and the ground truth of 3D geometry poses, and the absolute query pose can be naturally calculated. Furthermore, Relocnet \cite{balntas2018relocnet} introduces a frustum overlap loss to assist global descriptors learning that are suitable for camera localization. Motivated by these, CamNet \cite{ding2019camnet} applies two stages retrieval, image-based coarse retrieval and pose-based fine retrieval, to select the most similar reference frames for finally precise pose estimation. Without the need of training on specific scenarios, reference-based approaches are naturally scalable and flexible to be utilized in new scenarios. 
Since reference-based methods need to maintain a database of geo-tagged images, they are more trivial to scale to large-scale scenarios, compared with the structure-based counterparts. Overall, these image retrieval based methods achieve a trade-off between accuracy and scalability.


\subsubsection{Implicit Map Based Localization}
Implicit map based localization directly regresses camera pose from single images, by implicitly representing the structure of entire scene inside a deep neural network. 
The common pipeline is illustrated in Figure \ref{fig: 2d2d} (b) - the input to a neural network is single images, while the output is the global position and orientation of query images.

PoseNet \cite{kendall2015posenet} is the first work to tackle camera relocalization problem by training a ConvNet to predict camera pose from single RGB images in an end-to-end manner. PoseNet is based on the main structure of GoogleNet \cite{szegedy2015going} to extract visual features,, but removes the last softmax layers. Instead, a fully connected layer was introduced to output a 7 dimensional global pose, consisting of position and orientation vector in 3 and 4 dimensions respectively. However, PoseNet was designed with a naive regression loss function without any consideration for geometry, in which the hyper-parameters inside requires expensive hand-engineering to be tuned. Furthermore, it also suffers from the overfitting problem due to the high dimensionality of the feature embedding and limited training data. Thus various extensions enhance the original pipeline by exploiting LSTM units to reduce the dimensionality \cite{walch2017image}, applying synthetic generation to augment training data \cite{naseer2017deep, wu2017delving, purkait2018synthetic}, replacing backbone with ResNet34 \cite{melekhov2017image}, modelling pose uncertainty \cite{kendall2016modelling, cai2018hybrid} and introducing geometry-aware loss function \cite{kendall2017geometric}. Alternatively, Atloc \cite{wang2019atloc} associates the features in spatial domain with attention mechanism, which encourages the network to focus on parts of the image that are temporally consistent and robust. Similarly, a prior guided dropout mask is additionally adopted in RVL \cite{huang2019prior} to further eliminate the uncertainty caused by dynamic objects. Different such methods only considering spatial connections, VidLoc \cite{Clark2017} incorporates temporal constraints of image sequences to model the temporal connections of input images for visual localization. Moreover, additional motion constraints, including spatial constraints and other sensor constraints from GPS or SLAM systems are exploited in MapNet \cite{Brahmbhatt2018}, to enforce the motion consistency between predicted poses. Similar motion constraints are also added by jointly optimizing a relocalization network and visual odometry network \cite{valada2018deep, xue2019local}. However, being application-specific, scene representations learned from localization tasks may ignore some useful features they are not designed for. Out of this, VLocNet++ \cite{radwan2018vlocnet++} and FGSN \cite{larsson2019fine} additionally exploits the inter-task relationship among learning semantics and regressing poses, achieving impressive results. 

Implicit map based localization approaches take the advantages of deep learning in automatically extracting features, that play a vital role in global localization in featureless environments, where conventional methods are prone to fail. However, the requirement of scene-specific training prohibits it from generalizing to unseen scenes without being retrained. Also, current implicit map based approaches have not shown comparable performance over other explicit map based methods \cite{sattler2019understanding}.

\subsection{2D-to-3D Localization}
\begin{figure}
    \centering
    \includegraphics[width=0.4\textwidth]{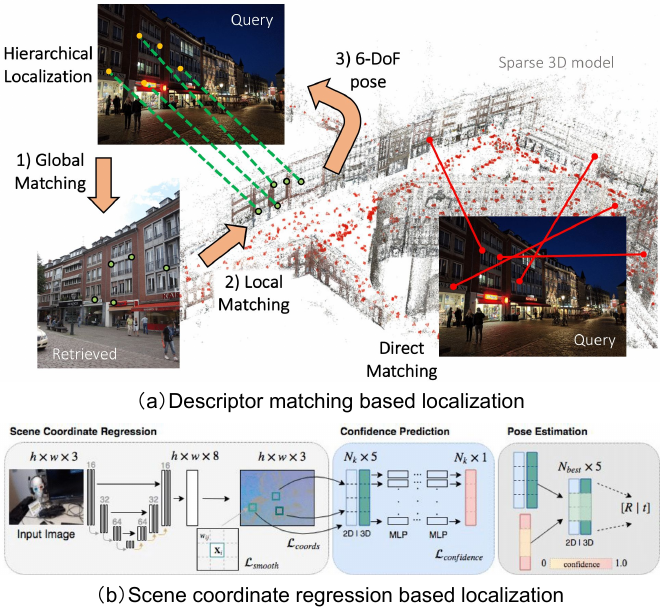}
    \caption{The typical architectures of 2D-to-3D based localization through (a) descriptor matching, i.e. HF-Net \cite{sarlin2019coarse} and (b) scene coordinate regression, i.e. Confidence SCR \cite{bui2018scene}.}
    \label{fig: structure}
\end{figure}

2D-to-3D localization refers to methods that recover the camera pose of a 2D image with respect to a 3D scene model. This 3D map is pre-built before performing global localization, via approaches such as structure from motion (SfM)\cite{saputra2018visual}. 
As shown in Figure \ref{fig: structure}, 2D-to-3D approaches establish 2D-3D correspondences between the 2D pixels of query image and the 3D points of scene model through local descriptor matching \cite{li2010location, li2012worldwide, zeisl2015camera} or by regressing 3D coordinates from pixel patches \cite{cavallari2017fly, guzman2014multi, shotton2013scene, massiceti2017random}. Such 2D-3D matches are then used to calculate camera pose by applying a Perspective-n-Point (PnP) solver \cite{gao2003complete, lepetit2009epnp} inside a RANSAC loop \cite{fischler1981random}. 

\begin{table*}[t]
\caption{A summary on existing methods on deep learning for 2D-to-3D global localization}
\label{tb: learning for 2d-3d localization}
    \small
    \begin{center}
\begin{tabular}{c c l c c c l}
\hline
\multicolumn{3}{c}{\multirow{2}{*}{Model}} & \multirow{2}{*}{Agnostic} & \multicolumn{2}{c}{Performance (m/degree)} & \multicolumn{1}{c}{\multirow{2}{*}{Contributions}} \\
\multicolumn{3}{c}{} & & 7Scenes & Cambridge & \multicolumn{1}{c}{}\\ \hline
\multirow{29}{*}{\rotatebox{90}{2D-3D Localization}}
& \multirow{18}{*}{\rotatebox{90}{Descriptor Based}}                            
& NetVLAD \cite{arandjelovic2016netvlad} & Yes & -  & -  & differentiable VLAD layer\\
&& DELF \cite{noh2017large} & Yes & - & - & attentive local feature descriptor\\
&& InLoc \cite{taira2018inloc} & Yes & 0.04/1.38  & 0.31/0.73 & dense data association \\
&& SVL \cite{schonberger2018semantic} & No  & - & - & leverage a generative model for descriptor learning \\ 
&& SuperPoint \cite{detone2018superpoint} & Yes & - & - & jointly extract interest points and descriptors \\ 
&& Sarlin et al. \cite{sarlin2018leveraging} & Yes & - & - & hierarchical localization \\ 
&& NC-Net \cite{rocco2018neighbourhood} & Yes & - & - & neighbourhood consensus constraints \\
&& 2D3D-MatchNet \cite{feng20192d3d} & Yes & - & - & jointly learn the descriptors for 2D and 3D keypoints\\
&& Unsuperpoint \cite{christiansen2019unsuperpoint} & Yes & - & - & unsupervised detector and descriptor learning\\
&& HF-Net \cite{sarlin2019coarse} & Yes & - & - & coarse-to-fine localization \\ 
&& D2-Net \cite{dusmanu2019d2} & Yes & - & - & jointly learn keypoints and descriptors \\
&& Speciale et al \cite{speciale2019privacy} & No  & - & - & privacy preserving localization\\
&& OOI-Net \cite{weinzaepfel2019visual} & No  & - & - & objects-of-interest annotations \\
&& Camposeco et al. \cite{camposeco2019hybrid} & Yes & - & 0.56/0.66 & hybrid scene compression for localization\\
&& Cheng et al. \cite{cheng2019cascaded} & Yes & - & - & cascaded parallel filtering \\
&& Taira et al. \cite{taira2019right} & Yes & - & - & comprehensive analysis of pose verification \\
&& R2D2 \cite{revaud2019r2d2} & Yes & - & - & learn a predictor of the descriptor discriminativeness \\
&& ASLFeat \cite{luo2020aslfeat} & Yes & - & - & leverage deformable convolutional networks\\ \cline{2-7}
& \multirow{11}{*}{\rotatebox{90}{Scene Coordinate Regression}}                 
& DSAC \cite{brachmann2017dsac} & No  & 0.20/6.3   & 0.32/0.78 & differentiable RANSAC \\
&& DSAC++ \cite{brachmann2018learning} & No  & 0.08/2.40  & 0.19/0.50 & without using a 3D model of the scene\\
&& Angle DSAC++ \cite{li2018scene} & No  & 0.06/1.47  & 0.17/0.50 & angle-based reprojection loss \\
&& Dense SCR \cite{li2018full} & No  & 0.04/1.4 & -  & full frame scene coordinate regression\\
&& Confidence SCR \cite{bui2018scene} & No  & 0.06/3.1 & - & model uncertainty of correspondences\\
&& ESAC \cite{brachmann2019expert} & No  & 0.034/1.50 & - & integrates DSAC in a Mixture of Experts\\
&& NG-RANSAC \cite{brachmann2019neural} & No & - & 0.24/0.30 & prior-guided model hypothesis search\\
&& SANet \cite{yang2019sanet} & Yes & 0.05/1.68  & 0.23/0.53 & scene agnostic architecture for camera localization \\
&& MV-SCR \cite{cai2019camera} & No  & 0.05/1.63  & 0.17/0.40 & multi-view constraints\\
&& HSC-Net \cite{li2020hscnet} & No  & 0.03/0.90  & 0.13/0.30 & hierarchical scene coordinate network\\
&& KFNet \cite{zhou2020kfnet} & No  & 0.03/0.88  & 0.13/0.30 & extends the problem to the time domain
\\ \hline
\end{tabular}
\end{center}
\begin{itemize}
              \footnotesize{
                \item \textit{Agnostic} indicates whether it can generalize to new scenarios.
                \item \textit{Performance} reports the position (m) and orientation (degree) error (a small number is better) on the 7-Scenes (Indoor)\cite{shotton2013scene} and Cambridge (Outdoor) dataset\cite{kendall2015posenet}. Both datasets are split into training and testing set. We report the averaged error on the testing set. 
                \item \textit{Contributions} summarize the main contributions of each work compared with previous research.
            }
    \end{itemize}
\end{table*}

\subsubsection{Descriptor Matching Based Localization}
\label{section:Descriptor Based}
Descriptor matching methods mainly rely on feature detector and descriptor, and establish the correspondences between the features from 2D input and 3D model. They can be further divided into three types: detect-then-describe, detect-and-describe, and describe-to-detect, according to the role of detector and descriptor in the learning process. 

\textit{Detect-then-describe} approach first performs feature detection and then extracts a feature descriptor from a patch centered around each keypoint \cite{mikolajczyk2004scale, leutenegger2011brisk}. The keypoint detector is typically responsible for providing robustness or invariance against possible real issues such as scale transformation, rotation, or viewpoint changes by normalizing the patch accordingly. However, some of these responsibilities might also be delegated to the descriptor. The common pipeline varies from using hand-crafted detectors \cite{bay2006surf, lowe2004distinctive} and descriptors \cite{calonder2010brief, rublee2011orb}, replacing either the descriptor \cite{balntas2016learning, simo2015discriminative, simonyan2014learning, rocco2018neighbourhood, moo2018learning, ebel2019beyond} or detector \cite{savinov2017quad, zhang2018learning, laguna2019key} with a learned alternative, or learning both the detector and descriptor \cite{ono2018lf, yi2016lift}. For efficiency, the feature detector often considers only small image regions and typically focuses on low-level structures such as corners or blobs \cite{harris1988combined}. The descriptor then captures higher level information in a larger patch around the keypoint. 

In contrast, \textit{detect-and-describe} approaches advance description stage. By sharing a representation from deep neural network, SuperPoint \cite{detone2018superpoint}, UnSuperPoint \cite{christiansen2019unsuperpoint} and R2D2 \cite{revaud2019r2d2} attempt to learn a dense feature descriptor and a feature detector. However, they rely on different decoder branches which are trained independently with specific losses. On the contrary, D2-net \cite{dusmanu2019d2} and ASLFeat \cite{luo2020aslfeat} shares all parameters between detection and description and uses a joint formulation that simultaneously optimizes for both tasks. 

Similarly, the \textit{describe-to-detect} approach, e.g. D2D \cite{tian2020d2d}, also postpones the detection to a later stage but applies such detector on pre-learned dense descriptors to extract a sparse set of keypoints and corresponding descriptors. Dense feature extraction foregoes the detection stage and performs the description stage densely across the whole image \cite{choy2016universal, fathy2018hierarchical, savinov2017matching, schonberger2018semantic}. In practice, this approach has shown to lead to better matching results than sparse feature matching, particularly under strong variations in illumination \cite{sattler2018benchmarking, zhou2016evaluating}. Different from these works, which purely rely on image features, 2D3D-MatchNet \cite{feng20192d3d} proposed to learn local descriptors that allow direct matching of key points across a 2D image and 3D point cloud. Similarly, LCD \cite{pham2019lcd} introduced a dual auto-encoder architecture to extract cross-domain local descriptors. However, they still require pre-defined 2D and 3D keypoints separately, which will result in poor matching results caused by inconsistent keypoint selection rules.

\begin{figure}
    \centering
    \includegraphics[width=0.4\textwidth]{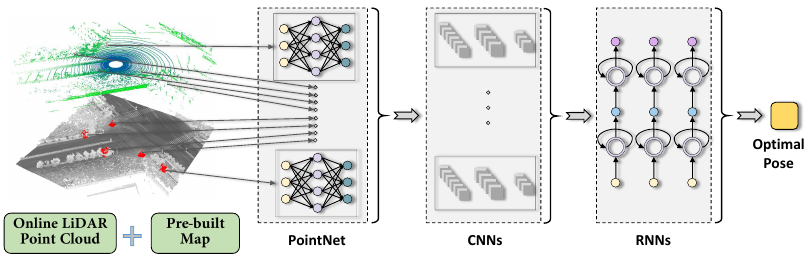}
    \caption{The typical architecture of 3D-to-3D localization, e.g. L3-Net \cite{lu2019l3}.}
    \label{fig: lidar}
\end{figure}

\begin{table*}
  		\caption{A summary of existing approaches on deep learning for 3D-to-3D localization} 
  		\label{tab: 3d-3d localization}
  		\small
  		\centering
  		\begin{tabular}{l c l}
  		\hline
    	Models	 & Agnostic & Contributions  \\
        \hline
        LocNet\cite{yin2018locnet} & No & convert 3D points into 2D matrix, search in global prior map  \\
        PointNetVLAD\cite{angelina2018pointnetvlad} & Yes & learn global descriptor from point clouds\\
        Barsan et al.\cite{barsan2018learning} & No & learn from LIDAR intensity maps and online point clouds\\
        L3-Net\cite{lu2019l3} & No & extract feature by PointNet\\
        PCAN\cite{zhang2019pcan} & Yes & predict the significance of each local point based on context\\
        DeepICP\cite{lu2019deepicp} & Yes & generate matching correspondence from learned matching probabilities\\
        DCP\cite{wang2019deep} & Yes & a learning based iterative closest point \\
        D3Feat\cite{bai2020d3feat} & Yes & jointly learn detector and descriptors for 3D points\\
        \hline
  		\end{tabular}
  		\begin{itemize}
              \footnotesize{
                \item \textit{Agnostic} indicates whether it can generalize to new scenarios.
                \item \textit{Contributions} summarize the main contributions of each work compared with previous research.
            }
    \end{itemize}
	\end{table*}

\subsubsection{Scene Coordinate Regression Based Localization}

Different from match-based methods that establish 2D-3D correspondences before calculating pose, 
scene coordinate regression approaches estimate the 3D coordinates of each pixel from the query image within the world coordinate system, i.e. the scene coordinates. It can be viewed as learning a transformation from the query image to the global coordinates of the scene. DSAC \cite{brachmann2017dsac} utilizes a ConvNet model to regress scene coordinates, followed by a novel differentiable RANSAC to allow end-to-end training of the whole pipeline. Such common pipeline was then improved by introducing the reprojection loss \cite{brachmann2018learning, brachmann2020visual, li2018scene} or multi-view geometric constraints \cite{cai2019camera} to enable unsupervised learning, jointly learning the observation confidences \cite{bui2018scene, brachmann2019neural} to enhance the sampling efficiency and accuracy, exploiting Mixture of Experts (MoE) strategy \cite{brachmann2019expert} or hierarchical coarse-to-fine \cite{li2020hscnet} to eliminate environment ambiguities. Different from these, KFNet \cite{zhou2020kfnet} extends the scene coordinate regression problem to the time domain and thus bridges the existing performance gap between temporal and one-shot relocalization approaches. However, they still trained for a specific scene and cannot be generalized to unseen scenes without retraining. To build a scene agnostic method, SANet \cite{yang2019sanet} regress the scene coordinate map of the query by interpolating the 3D points associated with retrieved scene images. Unlike aforementioned methods trained in a patch-based manner, Dense SCR \cite{li2018full} propose to perform the scene coordinate regression in a full-frame manner to make the computation efficient at test time and, more importantly, to add more global context to the regression process to improve the robustness. 

Scene coordinate regression methods often perform better robustness and higher accuracy under small indoor scenarios, outperforming traditional algorithms such as \cite{sattler2011fast}. But they have not yet proven their capacity in large-scale scenes.


\subsection{3D-to-3D Localization}
\label{subsec:LIDARLoc}

3D-to-3D localization (or LIDAR localization) refers to methods that recover the global pose of 3D points (i.e. LIDAR point cloud scans) against a pre-built 3D map by establishing a 3D-to-3D correspondence matching. Figure \ref{fig: lidar} shows the pipeline of 3D-to-3D localization: online scans or predicted coarse poses are applied to query the most similar 3D map data, which are further used for precise localization by calculating the offset between predicted poses and ground truths or estimating the relative poses between online scans and queried scene. 

By formulating LIDAR localization as a recursive Bayesian inference problem, \cite{barsan2018learning} embeds both LIDAR intensity maps and online point cloud sweeps in a sharing space for fully differentiable pose estimation. Instead of operating on 3D data directly, LocNet \cite{yin2018locnet} converts point cloud scans to 2D rotational invariant representation for searching similar frames in the global prior map, and then performs the iterative closest point (ICP) methods to calculate global pose. Towards proposing a learning based LIDAR localization framework that directly processes point clouds, L3-Net \cite{lu2019l3} processes point cloud data with PointNet \cite{qi2017pointnet} to extract feature descriptors that encode certain useful properties, and models the temporal connections of motion dynamics via a recurrent neural network. It optimizes the loss between the predicted poses and ground truth values by minimizing the matching distance between the point cloud input and the 3D map. Some techniques, such as PointNetVLAD \cite{angelina2018pointnetvlad}, PCAN \cite{zhang2019pcan} and D3Feat \cite{bai2020d3feat} explored to retrieve the reference scene at the beginning, while other techniques such as DeepICP \cite{lu2019deepicp} and DCP \cite{wang2019deep} allow to estimate relative motion transformations from 3D scans. 
Compared with image-based relocalization including 2D-to-3D and 2D-to-2D localization, 3D-to-3D localization is relatively underexplored.

\section{SLAM}
Simultaneously tracking self-motion and estimate the structure of surroundings constructs a simultaneous localization and mapping (SLAM) system. The individual modules of localization and mapping discussed in the above sections can be viewed as modules of a complete SLAM systems. 
This section overviews the SLAM systems using deep learning, with the main focus on the modules that contribute to the integration of a SLAM system, including local/global optimization, keyframe/loop closure detection and uncertainty estimation. Table \ref{tab: slam} summarizes the existing approaches that employ the deep learning based SLAM modules discussed in this section.

\subsection{Local Optimization}
When jointly optimizing estimated camera motion and scene geometry, SLAM systems enforce them to satisfy a certain constraint. This is done by minimizing a geometric or photometric loss to ensure their consistency in the local area - the surroundings of camera poses
, which can be viewed as a bundle adjustment (BA) problem\cite{triggs1999bundle}.
Learning based approaches predict depth maps and ego-motion through two individual networks \cite{Zhou2017} trained above large datasets. During the testing procedure when deployed online, there is a requirement that enforces the predictions to satisfy the local constraints. 
To enable local optimization, traditionally, the second-order solvers, e.g. Gauss-Newton (GN) method or Levenberg-Marquadt (LM) algorithm \cite{nocedal2006numerical}, are applied to optimize motion transformations and per-pixel depth maps


To this end, LS-Net \cite{clark2018learning} tackled this problem via a learning based optimizer by integrating the analytical solvers into its learning process. It learns a data-driven prior, followed by refining the DNN predictions with an analytical optimizer to ensure photometric consistency.
BA-Net \cite{tang2019ba} integrates a differentiable second-order optimizer (LM algorithm) into a deep neural network to achieve end-to-end learning. Instead of minimizing geometric or photometric error, BA-Net is performed on feature space to optimize the consistency loss of features from multiview images extracted by ConvNets. 
This feature-level optimizer can mitigate the fundamental problems of geometric or photometric solution, i.e. some information may be lost in the geometric optimization, while environmental dynamics and lighting changes may impact the photometric optimization). These learning based optimizers provide an alternative to solve bundle adjustment problem.

  \begin{table}
  		\caption{A summary of existing approaches on deep learning for SLAM} 
  		\label{tab: slam}
  		\small
  		\centering
  		\begin{tabular}{l l}
  		\hline
    	Modules	 & Employed by  \\
        \hline
        Local optimization & \cite{clark2018learning,tang2019ba} \\
        Global optimization & \cite{tateno2017cnn,zhou2020deeptam,li2019pose,czarnowski2020deepfactors} \\
        Keyframe detection & \cite{sheng2019unsupervised} \\
        Loop-closure detection & \cite{sunderhauf2015place,gao2017unsupervised,merrill2018lightweight,memon2020loop} \\
        Uncertainty Estimation & \cite{wang2018end,kendall2016modelling,Clark2017} \\
        \hline
  		\end{tabular}
	\end{table}

\subsection{Global Optimization}
Odometry estimation suffers from the accumulative error drifts during long-term operations, due to the fundamental problems of path integration, i.e. the system error accumulate without effective restrictions.
To address this, graph-SLAM \cite{grisetti2010tutorial} constructs a topological graph to represent camera poses or scene features as graph nodes, which are connected by edges (measured by sensors) to constrain the poses. 
This graph-based formulation can be optimized to ensure the global consistency of graph nodes and edges, mitigating the possible errors on pose estimates and the inherent sensor measurement noise. A popular solver for global optimization is through Levenberg-Marquardt (LM) algorithm.

In the era of deep learning, deep neural networks excel at extracting features, and constructing functions from observations to poses and scene representations. A global optimization upon the DNN predictions is necessary to reducing the drifts of global trajectories and support large-scale mapping.
Compared with a variety of well-researched solutions in classical SLAM, optimizing deep predictions globally is underexplored.

Existing works explored to combine learning modules into a classical SLAM system at different levels - in the front-end, DNNs produce predictions as priors, followed by incorporating these deep predictions into the back-end for next step optimization and refinement. One good example is CNN-SLAM \cite{tateno2017cnn}, which utilizes the learned per-pixel depths into LSD-SLAM \cite{engel2014lsd}, a full SLAM system to support loop closing and graph optimization. Camera poses and scene representations are jointly optimized with depth maps to produce consistent scale metrics.
In DeepTAM \cite{zhou2020deeptam}, both the depth and pose predictions from deep neural networks are introduced into a classical DTAM system \cite{Newcombe2011}, that is optimized globally by the back-end to achieve more accurate scene reconstruction and camera motion tracking. A similar work can be found on integrating unsupervised VO with a graph optimization back-end \cite{li2019pose}.
DeepFactors \cite{czarnowski2020deepfactors} vice versa integrates the learned optimizable scene representation (their so-called code representation) into a different style of back-end - probabilistic factor graph for global optimization. The advantage of the factor-graph based formulation is its flexibility to include sensor measurements, state estimates, and constraints. 
In a factor graph bach-end, it is quite easy and convenient to add new sensor modalities, pairwise constraints and system states into the graph for optimization.
However, these back-end optimizers are not yet differentiable.


\subsection{Keyframe and Loop-closure Detection}

Detecting keyframe and loop-closing is of key importance to the back-end optimization of SLAM systems. 

Keyframe selection facilitates SLAM systems to be more efficient. In the key-frame based SLAM systems, pose and scene estimates are only refined when a keyframe is detected. \cite{sheng2019unsupervised} provides a learning solution to detect key-frames together with unsupervised learning of ego-motion tracking and depths estimation \cite{Zhou2017}. Whether an image is the keyframe is determined by comparing its feature similarity with existing keyframes (i.e. if the similarity is below a threshold, this image will be treated as a new keyframe). 

Loop-closure detection or place recognition is also an important module in SLAM back-end to reduce open-loop errors. Conventional works are based on bag-of-words (BoW) to store and use the visual features from the hand-crafted detectors.
However, this problem is complicated by the changes of illumination, weather, viewpoints and moving objects in real-world scenarios. To solve this, previous researchers such as \cite{sunderhauf2015place} proposed to use the ConvNet features instead, that are from a pre-trained model on a generic large-scale image processing dataset. These methods are more robust against the variance of viewpoints and conditions due to the high-level representations extracted by deep neural networks.
Other representative works \cite{gao2017unsupervised,merrill2018lightweight,memon2020loop} are built on deep auto-encoder structure to extract a compact representation, that compresses scene in an unsupervised manner. 
Deep learning based loop closing contributes more robust and effective visual features, and achieves state-of-the-art performance in place recognition, which is suitable to be integrated in SLAM systems. 

\subsection{Uncertainty Estimation}

Safety and interpretability are an critical step towards the practical deployment of mobile agents in everyday life: the former enables agents to live and act with human reliably, while the latter allows users to have better understanding over the model behaviours. Although deep learning models achieve state-of-the-art performance in a wide range of regression and classification tasks, some corner cases should be given enough attention as well.
In these failure cases, errors from one component will propagate to the other downstream modules, causing catastrophic consequences. 
To this end, there is an emerging need to estimate uncertainty for deep neural networks to ensure safety and provide interpretability.

Deep learning models usually only produce the mean values of predictions, for example, the output of a DNN-based visual odometry model is a 6-dimensional relative pose vector, i.e. the translation and rotation. In order to capture the uncertainty of deep models, learning models can be augmented into a Bayesian model \cite{gal2016dropout,kendall2017uncertainties}. The uncertainty from Bayesian models is broadly categorized into Aleatoric uncertainty and epistemic uncertainty: Aleatoric uncertainty reflects observation noises, e.g. sensor measurement or motion noises; epistemic uncertainty captures the model uncertainty \cite{kendall2017uncertainties}. 
In the context of this survey, we focus on the work of estimating uncertainty on the specific task of localization and mapping, with regard to their usages, i.e. whether they capture the uncertainty with the purpose of motion tracking or scene understanding. 

The uncertainty of DNN-based odometry estimation has been explored by \cite{wang2018end,chen2019deep}.
They adopted a common strategy to convert the target predictions into a Gaussian distribution, conditioned on the mean value of pose estimates and its covariance. The parameters inside the framework are optimized via the loss function with a combination of mean and covariance. By minimizing the error function to find the best combination, the uncertainty is automatically learned in an unsupervised fashion.
In this way, the uncertainty of motion transformation is recovered. 
The motion uncertainty plays a vital role in probabilistic sensor fusion or the back-end optimization of SLAM systems. 
To validate the effectiveness of uncertainty estimation in SLAM systems,
\cite{wang2018end} integrated the learned uncertainty into a graph-SLAM as the covariances of odometry edges. Based on these covariances a global optimization is then performed to reduce system drifts. It also confirms that uncertainty estimation improves the performance of SLAM systems over the baseline with a fixed predefined value of covariance. 
Similar Bayesian models are applied to the global relocalization problem. As illustrated in \cite{kendall2016modelling,Clark2017}, the uncertainty from deep models are able to reflect the global location errors, in which the unreliable pose estimates are avoided with this belief metric. 

In addition to the uncertainty for motion/relocalization, estimating the uncertainty for scene understanding also contributes to SLAM systems. This uncertainty offers a belief metric in to what extent the environmental perception and scene structure should be trusted. For example, in the semantic segmentation and depth estimation tasks, uncertainty estimation provides per-pixel uncertainties for the DNN predictions \cite{kendall2017bayesian,kendall2017uncertainties,mcallister2017concrete,klodt2018supervising}. 
Further more, scene uncertainty is applicable to building hybrid SLAM systems. For example, photometric uncertainty can be learned to capture the variance of intensity on each image pixel, and hence enhances the robustness of SLAM system to observation noise \cite{yang2020d3vo}.

\section{Open Questions}

Although deep learning has brought great success to the research in localization and mapping, as aforementioned, existing models are not sophisticated enough to completely solve the problem at hand. The current form of deep solutions is still in its infancy state. Towards great autonomy in the wild, there are numerous challenges for future researchers to investigate. Practical applications of these techniques should be considered a systematic research problem. We discuss several open questions that likely lead the further development in this area.


    1) \textbf{End-to-end model vs. hybrid model}. End-to-end learning models are able to predict self-motion and scene directly from raw data, without any hand-engineering. Benefited from the advances of deep learning, end-to-end models are evolving fast to achieve increasing performance in accuracy, efficiency and robustness. Meanwhile, these models have been shown easier to be integrated with other high-level learning tasks, e.g. path planning and navigation\cite{zhu2017target}. Fundamentally, there exist underlying physical or geometric models to govern localization and mapping systems. Whether we should develop end-to-end models relying only on the power of data-driven approaches or integrate deep learning modules into the pre-built physical/geometric models as a hybrid model is a critical question for future research. As we can see, hybrid models already achieved the state-of-the art results in many tasks, e.g. visual odometry\cite{yang2020d3vo}, and global localization\cite{brachmann2018learning}. 
        Thus, it is reasonable to investigate in the way to take better advantage of the prior empirical knowledge from deep learning for hybrid models.
        On the other side, pure end-to-end models are data hunger. The performance of current models can be limited by the size of training dataset, and it is essential to create large and diverse dataset to enlarge the capacity of data-driven models.

     2) \textbf{Unifying evaluation benchmark and metric.} Finding suitable evaluating benchmark and metric is always a concern for SLAM systems. 
        This is especially the case for DNN based systems. The predictions from DNNs are affected by the characteristics of both training and test data, including the dataset size, hyperparameters (batch size and learning rate etc.), and the difference in testing scenarios. Therefore, it is hard to fairly compare them when considering dataset differences, training/testing configuration, or evaluation metric adopted in each work. 
        For example, the KITTI dataset is a common choice to evaluate visual odometry, but previous works split training and testing data in different ways (e.g. \cite{Wang2017,saputra2019learning,xue2019beyond} used Sequence 00, 02, 08, 09 as training set, and Sequence 03, 04, 05, 06, 07, 10 as testing set, while \cite{bian2019unsupervised,yang2020d3vo} used Sequence 00 - 08 as training, and left 09, and 10 as testing set). Some of them are even based on different evaluation metrics (e.g. \cite{Wang2017,saputra2019learning,xue2019beyond} applied the KITTI official evaluation metric, while \cite{Zhou2017,Zhan2018} applied absolute trajectory error (ATE) as their evaluation metric). All these factors bring difficulties to a direct and fair comparison across them. Moreover, the KITTI dataset is relatively simple (the vehicle only moves in 2D translation) and in small size. It is not convincing if only results on KITTI benchmark are provided without a comprehensive evaluation in the long-term real-world experiment. In fact, there is a growing need in creating a benchmark for a though system evaluation covering various environments, self-motions and dynamics. 
        
     3) \textbf{Real-world deployment}. Deploying deep learning models in real-world environments is a systematic research problem. 
        In the existing research, the prediction accuracy is always their `golden rule' to follow, while other crucial issues are overlooked, such as whether the model structure and the parameter number of framework is optimal. The computational and energy consumption have to be considered on resource-constrained systems, such as low-cost robots or VR wearable devices. The prallerization opportunities, such as convolutional filters or other parallel neural network modules should be exploited in order to take better use of GPUs. 
        Examples for consideration include in which situations the feed-back should be returned to fine-tune the systems, how to incorporate the self-supervised models into the systems and whether the systems allow the real-time online learning.
        
    4) \textbf{Lifelong learning}. Most previous works we discussed so far have only been validated on simple closed-form dataset
        , such as visual odometry and depth predictions are performed on the KITTI dataset. However, in an open world, the mobile agents will confront everchanging environmental factors, and moving dynamics. This will require the DNN models to continuously and coherently learn and adapt to the changes of the world. Moreover, new concepts and objects will appear unexpectedly, requiring an object discovery and new knowledge extension phase for robots.
        
    5) \textbf{New sensors}: Beyond the common choice of on-board sensors, such as cameras, IMU and LIDAR, the emerging new sensors provide an alternative to construct a more accurate and robust multimodal system. New sensors including event camera\cite{rebecq2016evo}, thermo camera\cite{saputra2020deeptio}, mm-wave device\cite{Chris2020}, radio signals\cite{ferris2007wifi}, magnetic sensor\cite{lu2018simultaneous}, have distinct properties and data format compared to predominant SLAM sensors such as cameras, IMU and LIDAR. Nevertheless, the effective learning approaches to processing these unusual sensors are still underexplored.

    6) \textbf{Scalability}. Both the learning based localization and mapping models have now achieved promising results on the evaluation benchmark. However, they are restricted to some scenarios. For example, odometry estimation is always evaluated in the city area or on the roads. Whether these techniques could be applied to other environments, e.g. rural area or forest area is still an open question. Moreover, existing works on scene reconstruction are restricted on single-objects, synthetic data or room level. It is worthy exploring whether these learning methods are capable of scaling to much more complex and large-scale reconstruction problems.

    7) \textbf{Safety, reliability and interpretability.} Safety and reliability are critical to practical applications, e.g. self-driving vehicles. In these scenarios, even a small error of pose or scene estimates will cause disasters to the entire system. Deep neural networks have been long-critisized as 'black-box', exacerbating the safety concerns for critical tasks. Some initial efforts explored the interpretability on deep models \cite{zhang2018visual}. For example, uncertainty estimation\cite{gal2016dropout,kendall2017uncertainties} can offer a belief metric, representing to what extent we trust our models. In this way, the unreliable predictions (with low uncertainty) are avoided in order to ensure the systems to stay safe and reliable. 



\section{Conclusions}

This work comprehensively overviews the area of deep learning for localization and mapping, and provides a new taxonomy to cover the relevant existing approaches from robotics, computer vision and machine learning communities. 
Learning models are incorporated into localization and mapping systems to connect input sensor data and target values, by automatically extracting useful features from raw data without any human effort. Deep learning based techniques have so far achieved the state-of-the-art performance in a variety of tasks, from visual odometry, global localization to dense scene reconstruction. Due to the highly expressive capacity of deep neural networks, these models are capable of implicitly modelling the factors such as environmental dynamics or sensor noises, that are hard to be modelled by hand, and thus are relatively more robust in real-world applications. In addition, high-level understanding and interaction are easy to perform for mobile agents with the learning based framework. 
The fast development of deep learning provides an alternative to solve classical localization and mapping problem in a data-driven way, and meanwhile paves the road towards a next-generation AI based spatial perception solution.


\ifCLASSOPTIONcompsoc
  \section*{Acknowledgments}
\else
  \section*{Acknowledgment}
\fi

This work is supported by the EPSRC Project ``ACE-OPS: From Autonomy to Cognitive assistance in Emergency OPerationS'' (Grant Number: EP/S030832/1).

\bibliography{reference}
\bibliographystyle{ieeetr}

\end{document}